\documentclass{article}

\usepackage{microtype}
\usepackage{graphicx}
\usepackage{subfigure}
\usepackage{booktabs} 

\usepackage{layouts}
\usepackage{enumerate}

\usepackage{hyperref}

\usepackage[preprint]{icml2026}

\usepackage{amsmath}
\usepackage{amssymb}
\usepackage{amsfonts}
\usepackage{mathtools}
\usepackage{amsthm}
\usepackage{graphicx}
\usepackage{wrapfig}
\usepackage{longtable,booktabs,makecell,array}
\usepackage{multirow}

\usepackage[capitalize,noabbrev]{cleveref}

\usepackage{amsmath,amsfonts,bm,  amsthm}

\def\eqref#1{equation~\ref{#1}}

\def\1{\bm{1}}

\def\vTheta{{\bm{\Theta}}}
\def\vM{{\bm{\mathcal{M}}}}
\def\vm{{\mathcal{M}}}
\def\vzero{{\bm{0}}}

\def\vtheta{{\bm{\theta}}}

\def\vf{{\bm{f}}}

\def\vn{{\bm{n}}}

\def\vs{{\bm{s}}}

\def\vx{{\bm{x}}}

\def\vz{{\bm{z}}}

\DeclareMathAlphabet{\mathsfit}{\encodingdefault}{\sfdefault}{m}{sl}
\SetMathAlphabet{\mathsfit}{bold}{\encodingdefault}{\sfdefault}{bx}{n}

\theoremstyle{plain}

\theoremstyle{definition}

\theoremstyle{remark}

\definecolor{myorange1}{RGB}{255, 102, 0}
\newcommand\myorange[1]{{\color{myorange1}{#1}}}

\definecolor{mygreen1}{RGB}{68, 170, 0}

\definecolor{myblue1}{RGB}{0, 178, 255}
\newcommand\myblue[1]{{\color{myblue1}{#1}}}

\usepackage[textsize=tiny]{todonotes}

\newcommand{\ours}{PRISM}
\icmltitlerunning{Scalable Simulation-Based Model Inference with Test-Time Complexity Control}

\begin{document}

\twocolumn[
\icmltitle{
Scalable Simulation-Based Model Inference with Test-Time Complexity Control
}



\icmlsetsymbol{equal}{*}

\begin{icmlauthorlist}
\icmlauthor{Manuel Gloeckler}{tue}
\icmlauthor{J.P. Manzano-Patrón}{uk}
\icmlauthor{Stamatios N. Sotiropoulos}{uk,uk2}
\icmlauthor{Cornelius Schröder}{tue}
\icmlauthor{Jakob H. Macke}{tue,mpi}

\end{icmlauthorlist}

\icmlaffiliation{tue}{Machine Learning in Science, University of Tübingen and Tübingen AI Center, Tübingen, Germany}
\icmlaffiliation{mpi}{Max Planck Institute for Intelligent Systems, Department Empirical Inference, Tübingen, Germany}
\icmlaffiliation{uk}{Sir Peter Mansfield Imaging Centre, School of Medicine, University of Nottingham, UK}
\icmlaffiliation{uk2}{National Institute for Health Research (NIHR) Nottingham Biomedical Research Centre, Queens Medical Centre, Nottingham, United
Kingdom}

\icmlcorrespondingauthor{Manuel Gloeckler}{manuel.gloeckler@uni-tuebingen.de}
\icmlcorrespondingauthor{Jakob H. Macke}{jakob.macke@uni-tuebingen.de}

\icmlkeywords{Machine Learning, ICML}

\vskip 0.3in
]



\printAffiliationsAndNotice{\icmlEqualContribution} 

\begin{abstract}
Simulation plays a central role in scientific discovery. In many applications, the bottleneck is no longer running a simulator---it is choosing among large families of plausible simulators, each corresponding to different forward models/hypotheses consistent with observations. 
Over large model families, classical Bayesian workflows for model-selection 
are impractical. Furthermore, \emph{amortized} model-selection methods typically hard-code a fixed model prior---or complexity penalty---at training time, requiring users to commit to a particular parsimony assumption before seeing the data.
We introduce \emph{\ours}, a simulation-based encoder-decoder that infers a joint posterior over both discrete model structures and associated continuous parameters, while enabling test-time control of model complexity via a tunable model prior that the network is conditioned on.
We show that \ours~ scales to families with combinatorially many (up to  billions) of  model instantiations on a synthetic symbolic regression task. 
As a scientific application, we evaluate \ours\ on biophysical modeling for diffusion MRI data, showing the ability to perform model selection across several multi-compartment models, on both synthetic and in-vivo neuroimaging data. 
\end{abstract}
\vspace{-0.5cm}

\begin{figure}[tp]
    \centering
    \vspace{-0cm}
    \includegraphics[width=0.95\columnwidth]
    {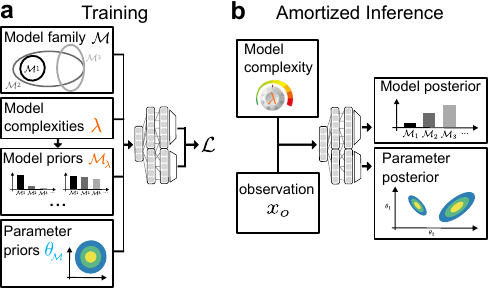}
    \vspace{-0.0cm}
    \caption{\textbf{\ours~ overview. (a)}
    During training, we sample from a model family $\vm$ with a hierarchical prior $p(\mathcal{M}\mid\lambda)$, where $\lambda$ controls a chosen notion of model complexity and approximate jointly the model posterior and the conditional parameter posterior $p(\mathcal{M},\theta \mid \vx, \lambda)$ by optimizing the loss $\mathcal{L}=\mathcal{L}(\myorange{\vm}, p(\vm \mid \vx, \myorange{\lambda} )) + \mathcal{L}(\myblue{\theta_\vm} , p(\theta \mid \myorange{\vm_\lambda}, \vx))$, for model evaluations $\vx\sim p_\myorange{\vm}(\vx \mid \myblue{\theta_\vm} )$.
    \textbf{(b)} At inference time, we set $\lambda$ to tune parsimony, and select or explore models in the combinatorial space, infer model parameters, and deploy the resulting posteriors in downstream analyses.}
    \label{fig:illustration}
    \vspace{-0.55cm}
\end{figure}

\section{Introduction}
\label{sec:intro}

Simulation is a cornerstone of scientific inquiry and discovery ~\citep{winsberg2019scienceintheageofsimulation,skuse2019third} and is increasingly relevant as researchers tackle more complex scientific systems~\citep{lavin2022simulationintelligencenewgeneration}. In practice, researchers often face multiple plausible models for the same process, reflecting different mechanistic assumptions, parameterizations, and levels of abstraction. This proliferation creates a persistent challenge across modeling communities: deciding which model configurations are most supported by data in a given experimental regime, and how strongly.

For example, in computational neuroscience, Hodgkin– Huxley-type models~\citep{hodgkin1952quantitative} can be composed of many candidate ion-channel and morphological components \cite{schoelzel2020modularHH} to represent neural activity and propagation. In epidemiology, the canonical SIR model~\citep{kermack1927contribution} has spawned a broad family of compartmental extensions (e.g., SEIR/SIRS, stratified or structured variants), with trade-offs between interpretability, identifiability, and predictive performance~\citep{hethcote2000mathematics,keeling2008modeling}. In diffusion magnetic resonance imaging (dMRI), a range of multi-compartment biophysical models have been proposed as potential biomarkers to indirectly infer tissue microstructure from the scatter pattern of water molecules within the brain \citep{basser1994dti,jelescu2020challenges}.
Symbolic regression \cite{brunton2016discovering_SINDY, biggio2021neuralsymbolicregreassionthatscales} can be seen through the same lens, where equations are constructed by selecting components from a large library.
In all these cases, models formalize competing hypotheses, and selecting among them is central to scientific discovery. Yet, even modest libraries of components induce combinatorial  model spaces that cannot be exhaustively enumerated or fit one by one. The central task is to infer which component combinations and parameters are supported by data, often prioritizing simple explanations over complex ones.


Bayesian model selection offers a principled framework to address this challenge, in particular by using criteria based on model-evidences (marginal likelihoods).  However, there are two central challenges: First, Bayesian model selection methods must operate reliably over large model families. Conventional workflows \citep{gelman2020bayesian} are based on running Bayesian inference separately for each candidate model (e.g., via MCMC) and comparing models via marginal likelihoods, which scale poorly with the number of models and require tractable likelihoods. A flexible alternative is to \emph{amortize} these costs by training neural networks to approximate key Bayesian computations once and reuse them across datasets. This connects to simulation-based inference (SBI) methods that learn posteriors, likelihood(-ratios) or model evidences from simulated data when likelihoods are unavailable \citep{cranmer2020frontier,radev2023bayesflowamortizedbayesianworkflows, deistler2025simulation_practicalguide}, and to “all-in-one” approaches that jointly learn posteriors and likelihood surrogates to support downstream model comparison \citep{radev2023jana,gloeckler2024allinonesimulationbasedinference, chang2025amortizedprobabilisticconditioningoptimization}.
A second key challenge is controlling the trade-off between model complexity and goodness-of-fit. In principle, a prior over models can be parameterized (here, by $\lambda$) to interpolate between “narrow” priors favoring simple models and “broad” priors allowing more complex ones. Choosing $\lambda$ in advance is difficult, and practitioners often adjust this parsimony trade-off after seeing the data. However, most model-selection or discovery methods effectively lock in these assumptions at training time, making post-hoc adjustment difficult without retraining.

\begin{figure}[tp]
    \centering
    \includegraphics[width=0.95\linewidth]{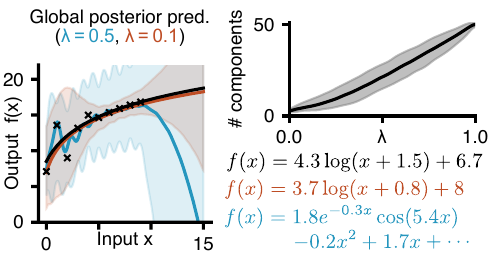}
    \vspace{-0.2cm}
    \caption{\textbf{Illustration  on symbolic regression task.} \emph{Left:} Ground-truth function and noisy observations (black, $x<10$), compared to posterior predictives samples for two  model complexities $\lambda$ (95\% credible interval and one noiseless sample each, chosen as the simplest posterior draw).
        \emph{Right:} Median number of model components as function of $\lambda$. 
        \emph{Below:} Equations of true function (black, 2 components), and two sampled equations for low and high complexity priors (2 and 9 components). 
    }
    \label{fig:symbolic_illustrative}
    \vspace{-0.45cm}
\end{figure}

We propose \emph{\ours}, a \emph{PRI}or-flexible \emph{S}imulation-based \emph{M}odel inference framework, based on a transformer encoder-decoder approach for simultaneous inference of models and model parameters (Fig.~\ref{fig:illustration}), which addresses both of these challenges. {\ours} can identify posteriors over both model-components and associated parameters even in large model-spaces. In addition, it can amortize over a test-time tunable model prior. This allows practitioners to explore different complexities (e.g., in symbolic regression expressions with different number of terms, Fig. \ref{fig:symbolic_illustrative})  and select one without having to retrain the pipeline. 
Compared to previous model-inference approaches \citep{schroder2024simultaneous}, we remove several highly restrictive design choices, e.g.,  the reliance on mixture density networks (and its associated need for analytic marginalization).   
Empirically, we show that \ours~ amortizes effectively over large model families with billions of distinct model candidates on a synthetic symbolic regression example. We then showcase the application of \ours~ to inference of tissue microstructure from  dMRI data, by considering a range of biophysical multi-compartment models (e.g. Ball-and-Sticks \cite{behrens2007multifibre}). We demonstrate that \ours~ can enable model comparison and selection in the context of fiber orientation estimation —on both synthetic and in-vivo data— and show that it outperforms previous SBI pipelines that consider each candidate model independently \citep{manzano2024uncertainty}. 

\section{Background and Notation}
\subsection{Bayesian parameter inference}
Classically, we start with a probabilistic model $\vm$ defined by model-specific parameters $\vtheta_\vm$ with prior $p(\vtheta_\vm\mid \vm)$ and a likelihood $p(\vx \mid \vtheta_\vm,\vm)$. Given model $\vm$ and data $\vx_o$, Bayes’ rule gives $p(\vtheta_\vm \mid \vx_o,\vm)\propto p(\vx_o \mid \vtheta_\vm,\vm)\,p(\vtheta_\vm\mid \vm)$. Classical methods (MCMC/VI) typically require model-specific implementations and repeated computation for inference per dataset $\vx_o$ \citep{gilks1995markov,beal2003}.

Simulation-based (a.k.a ``likelihood-free'') approaches purely operate on samples $\vx \sim p(\vx \mid \vtheta_\vm,\vm)$ without requiring likelihood-evaluations. They also readily permit \emph{amortized} inference, which trains a parameterized estimator (e.g., a neural posterior, likelihood, or likelihood-ratio estimator) on simulated pairs $(\vtheta_\vm,\vx)$ to enable fast test-time inference \citep{papamakarios2016fast,lueckmann2017flexible,radev2020bayesflow,radev2021amortized, jeffrey2024evidence, reuter2025can}.
However, standard SBI approaches generally assume a fixed, well-specified model; under mis-specification they may fail sharply out of distribution \citep{cannon2022investigating,kelly2025simulation}, making model construction and selection prerequisites for reliable amortization.

\subsection{Bayesian model selection}
To select an appropriate model, we consider a family of candidate simulators or probabilistic models ${\vM=\{\vm_1,\vm_2,\dots\}}$.
Bayesian model comparison treats the model identity $\vm \in \vM$ as a discrete latent variable
with prior $p(\vm)$. Conditioning on data $\vx_o$ gives
$P(\vm \mid \vx_o) \propto p(\vx_o \mid \vm)\,P(\vm)$. Therefore, the \emph{marginal likelihood} (Bayesian evidence)
$$p(\vx_o \mid \vm) = \int p(\vx_o \mid \vtheta_\vm,\vm)\,p(\vtheta_\vm \mid \vm)\,d\vtheta_\vm$$
is at the core of classical Bayesian model comparison
\citep[BMC]{Jeffreys1939TheoryProbability,kass1995bayesFactors,mackay2003informationTheory}. By averaging over the parameter prior, the evidence induces an Occam-type tradeoff: complex models that distribute
probability mass broadly are penalized relative to simpler models that concentrate mass near data  $\vx_o$ \citep{Jeffreys1939TheoryProbability,mackay2003informationTheory}. An alternative to model comparison or selection is the \textit{Bayesian model average}  over $P(\vm \mid \vx_o)$, which explicitly accounts for uncertainty over the model \citep{draper1995assessment}.

Computing the evidence requires high-dimensional integration, and is generally intractable. Furthermore, for large model families $\vM$, workflows that run inference and evidence estimation \emph{per model} become  prohibitive.  
While sometimes Bayesian model comparison is seen as nearly all-encompassing solution to model selection \citep{mackay2003ch28, lotfi2022bayesian} the marginal likelihood  penalizes a specific notion of \textit{complexity} implicitly defined by the prior, which may be misaligned with a  given task.

\section{Methods}
\label{sec:method}

\subsection{Problem setting}
We aim to estimate the joint posterior both over models and parameters $p(\vm, \vtheta | \vx_o, \lambda) = P(\vm|\lambda, \vx_o) p(\vtheta | \vm, \vx_o)$ for data $\vx \in \mathbb{R}^{d_x}$ (Fig.~\ref{fig:illustration}), as in \citet{schroder2024simultaneous}. To control \emph{model complexity} at inference time, we introduce a hierarchical prior on models $p(\mathcal{M}|\lambda),$ where $\lambda$ is a hyperparameter that controls model complexity. 
As model space $\vM$ we consider all combinations of $C$ potential components, and index individual models as binary mask $\mathcal{M} = (M_1,\dots, M_C)$ with $M_i \in \{0,1\}$, representing the presence or absence of a specific component. 
Each component is associated with either independent or shared parameters  $\vtheta_\vm \in \mathbb{R}^{d_\mathcal{M}}$ sampled from a model-specific parameter prior $ p(\vtheta_\mathcal{M}|\mathcal{M})$.

This setup yields two coupled approximation problems:
(i) learning the \emph{model posterior} $P(\mathcal{M}\mid \vx_o,\lambda)$ (represented as a  high-dimensional
multivariate binary distribution) and
(ii) learning the \emph{parameter posterior} $p(\vtheta_{\mathcal{M}}\mid \mathcal{M}_\lambda,\vx_o)$, whose
dimension varies with $\mathcal{M}$ (i.e., $\vtheta_\vm \subseteq \vTheta \in \mathbb{R}^{d_{\text{max}}}$, $d_{\mathcal{M}} < d_{\text{max}}$).
Our goal is to amortize both quantities over observation and model space ($\vx_o$, $\vM$), while allowing test-time adjustment of $\lambda$.

\begin{figure}
    \centering
    \includegraphics[width=\linewidth]{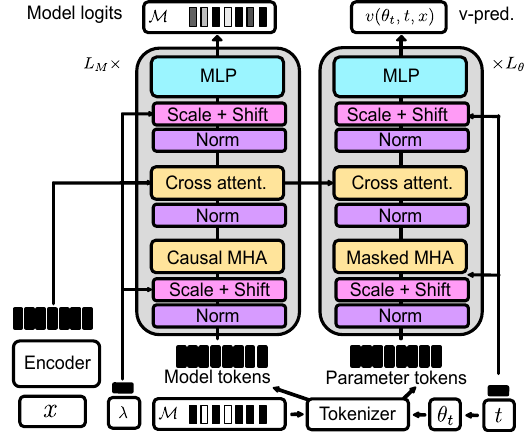}
    \caption{\textbf{Architecture overview.} \ours~ is based on two parallel transformers to infer (i) the model posterior from tokenized model masks and (ii) the parameter posterior via a diffusion $v$-prediction network ($t$ denotes diffusion time). Observations $\vx$ enter through cross-attention; $(\lambda, t)$ are injected via adaptive layer normalization (skip connections omitted). }
    \vspace{-0.2cm}
    \label{fig:architecture}
\end{figure}

\subsection{Model architecture of \ours}
\ours\ uses a transformer-based encoder-decoder architecture with two decoding streams: One for the discrete model structure and a second that targets a posterior over continuous parameters given a model structure. These decoders are paired with a task-specific tokenizer, and both conditioned on observations $\vx$ and model complexity $\lambda$  (Fig.~\ref{fig:architecture}).

\textbf{Model posterior decoder.}
We represent the model posterior as an autoregressive multivariate Bernoulli distribution over binary components $\mathcal{M}\in\{0,1\}^{C}$:
${q_\phi(\mathcal{M}\mid \vx,\lambda)
=\prod_{i=1}^{C} \mathrm{Bern}\!\Big(M_i ; p_\phi\!\left(M_{<i}, \vx,\lambda\right)\Big)}.$
We implement conditioning on $\vx$ via cross-attention to the encoder tokens, and inject the model-complexity parameter $\lambda$ through adaptive layer normalization (AdaLN,  \citealt{peebles2023scalable}; Fig.~\ref{fig:architecture}, $\lambda$-dependent scale and shift). A causal attention mask  with a padding token enables strict autoregressive decoding of the components.

\textbf{Parameter posterior decoder.}
To capture complex, potentially multimodal posteriors over continuous parameters, we use a diffusion-based decoder. We adopt the EDM noise parameterization (\citet{karras2022elucidatingdesignspacediffusionbased}; i.e. $\vtheta_t = \vtheta + t\,\boldsymbol{\epsilon}$, $ \boldsymbol{\epsilon}\sim\mathcal{N}(\mathbf{0},\mathbb{I})$),
and condition on the encoder context via cross-attention. Diffusion time $t$ is injected through AdaLN. The network is preconditioned for training on the $v$-prediction target (details in Sec.~\ref{app:model_training_details}),
\begin{equation*}
\mathbf{v}_t = \alpha_t \boldsymbol{\epsilon} - \beta_t \vtheta,
\quad
\mathbf{v}_\phi(\vtheta_t,t,\vx)=\alpha_t\,\hat{\boldsymbol{\epsilon}}_{\phi,\vtheta_t,t,\vx}-\beta_t\,\hat{\vtheta}_{\phi,\vtheta_t,t,\vx}.
\end{equation*}
The diffusion decoder defines an amortized approximation $q_\phi(\vtheta \mid \mathcal{M}, \vx, \lambda)$ over a global parameter vector $\vTheta\in\mathbb{R}^{d_{\max}}$. We condition on $\mathcal{M}$ using an $\mathcal{M}$-dependent \emph{block attention mask}: for any inactive component ($M_i=0$), we mask (or remove) the corresponding parameter tokens from the diffusion-transformer input by masking all attention connections to and from that component's parameter block (equivalently, remove the associated rows/columns in the attention matrix; Fig.~\ref{fig:architecture}, Masked MHA). This effectively marginalizes unused parameters, reducing $\vTheta$ to $\vtheta_{\mathcal{M}}$, while retaining a single shared parameter decoder across all models. When multiple components share global parameters, the associated tokens remain unmasked and attend to all active component tokens.

\begin{figure}[tp]
    \centering
    \includegraphics[width=\linewidth]{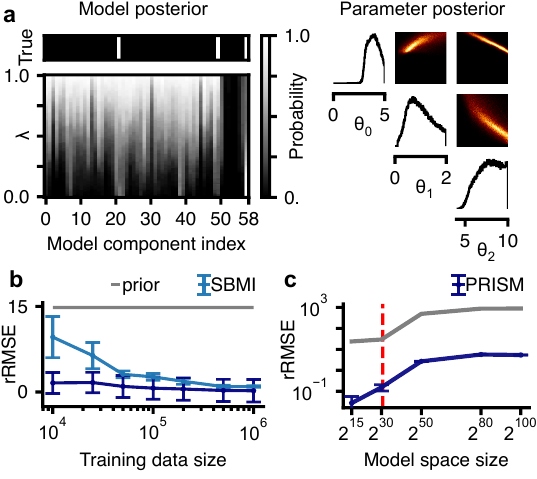}
    \caption{ \textbf{\ours~ on symbolic regression task. }
    \textbf{(a)} Model posterior across $\lambda$ and parameter posterior for the example in Fig.~\ref{fig:symbolic_illustrative} (for $\lambda = 0.1$).
    \textbf{(b)} Comparison to \citep{schroder2024simultaneous} for a fixed prior.
    \textbf{(c)} Scaling to large model spaces for a fixed computational training budget;  Red line indicates regimes beyond which not all models can be sampled during training.
    }
    \label{fig:symbolic_res}
\end{figure}

\textbf{Tokenizer.}
The tokenizer produces a token sequence for each decoder. For model components, each component  $i\in\{1,\dots,C\}$ is assigned a learnable component identification token, shared by both decoders. The model decoder additionally requires the binary mask $\mathcal{M}=(M_1,\dots,M_C)$; we encode this by adding a learned embedding of $M_i\in\{0,1\}$ to the corresponding token. The parameter decoder requires the diffusion-state input $\vtheta_t$ to be encoded as part of its token sequence. Rather than allocating a token per scalar parameter (as in autoregressive parameterizations), diffusion permits structured grouping of parameters. We therefore use per-component parameter tokens: each component $i$ is associated with a parameter subvector $\vtheta_i\in\mathbb{R}^{d_i}$, projected into the token space using a component-specific linear map. Components with $d_i=0$ contribute no parameter tokens, and shared parameters are represented by additional global tokens.

\textbf{Encoder.}
The encoder maps the observation $\vx$ (and optional auxiliary variables) to a sequence of fixed-size vectors. The encoder can be chosen in a task-dependent manner. We use a transformer encoder, with positional embeddings for sequential data and without positional embeddings for exchangeable sets (e.g., unordered measurements).
\subsection{Training}
Training requires joint samples $(\lambda, \mathcal{M},\vtheta_{\mathcal{M}},\vx)$ from the generative process. We optimize two losses jointly: (i) a Bernoulli negative log-likelihood for the model decoder (implemented as binary cross-entropy over $M_i$), and (ii) a diffusion $v$-prediction objective for the parameter decoder \citep{salimans2022progressive}. The overall training objective is the sum of these terms (Fig.~\ref{fig:illustration}, details in Appendix~\ref{app:model_training_details}). Since we target regimes with large $\mathcal{M}$-spaces, we use an {online } setup in which training data is generated on the fly. 
\subsection{Deployment}
Having both a \textit{model} and \textit{parameter} posterior enables uncertainty quantification at two levels: \emph{local} uncertainty conditional on a fixed model, and \emph{global} uncertainty that marginalizes over model uncertainty \citep{draper1995assessment, werner2021informed}. Furthermore, the model posterior enables Bayesian model comparison/selection via $q(\vm \mid \vx_o,\lambda)$.

\section{Results}
We apply \ours\ to two problems: We first illustrate and evaluate it on an additive symbolic regression task, before we apply it to dMRI to demonstrate its performance in a complex scientific application. 

\subsection{Toy Example:  Symbolic Regression}
\label{sec:symbolic_res}

We define the additive symbolic model family as 
$$ f(x| \vtheta, \mathcal{M}) = \sum_{k=0}^{K-1}  M_k \cdot g_k(x|\vtheta_k)  + \sum_{m=K}^{K + M-1} M_m \cdot  \epsilon_m(x|\vtheta_m) $$
where $g_k$ are ‘base functions’ and $\epsilon_m$ represent different noise models, with a total of  $C=K+M$ components.
We follow the protocol of \citet{schroder2024simultaneous}: each function is
evaluated on an equidistant grid over $[0,10]$,  
but we extend the basis library to $K=42$ unique base functions and $M=8$ noise models  (Tab. \ref{tab:extended_symbolic_components}). We allow repeated components, yielding up to $K=100$ components. Repeated components can have different parameters and thus induce distinct probabilistic models and create  posterior degeneracies that must be handled by the joint model--parameter inference networks. 
We define the complexity $\lambda$, and the corresponding family of model priors $\vm_\lambda$ as 
$p(\vm \mid \lambda) = \prod_k \mathrm{Ber}(M_k, \lambda) \prod_m \mathrm{Cat}(M_{K:M}),$
with $\lambda \sim \mathrm{Unif}([0,1])$. This definition gives rise to a simple yet intuitive complexity interpretation: $\lambda$ close to $1$ result in complex models with a lot of active model components $M_i =1$, whereas $\lambda \to 0$ results in sparse model samples. The noise models are unaffected and mutually exclusive. For all experiments we use as encoder a transformer with position encoding and 2 layers, and for model and parameter decoders 6 layers (details in Sec.~\ref{app:model_training_details}).

We first evaluate \ours~ at $K=50$ base functions. The model-averaged posterior predictive closely matches the observations in terms of RMSE (Fig.~\ref{fig:symbolic_illustrative}, \ref{fig:app:symbolic_illustrative_appendix}, \ref{fig:appendix_symbolic_metrics}, Sec.~\ref{app:subsec:eval_metrics}). Importantly, the posterior mass is spread across multiple distinct structures, indicating  model uncertainty (Fig.\ref{fig:symbolic_res}a).  Varying the complexity hyperparameter $\lambda$ reshapes the predicted posterior over $\mathcal{M}$, yielding structurally simpler explanations with smaller number of components (Fig.~\ref{fig:symbolic_illustrative}a, \ref{fig:symbolic_res}a). At small $\lambda$, the posterior concentrates on constant-offset components ($i=6,7,49,50$) and logarithmic ($i=20$) or logarithmic-like terms (e.g., $\sqrt{\cdot}$, $i=21$), consistent with the data (Tab.~\ref{tab:extended_symbolic_components}).
While the predictive fit is preserved within the training domain ($\vx <10$) deviations and uncertainty increases for more complex models for generalizations to $x>10$ (Fig.~\ref{fig:symbolic_illustrative} b).  
This illustrates that explicit test-time control of the model prior $p(\mathcal{M}\mid\lambda)$ enables selecting a desired level of sparsity without retraining.
%
For a direct comparison with prior work, we reproduce the largest setting in \citet{schroder2024simultaneous}
(SBMI, ${K=15}$) with fixed model prior. 
\ours~ clearly outperforms SBMI, even in the regime of small training data (Fig.~\ref{fig:symbolic_res}b).

\textbf{Scaling with model-space.}  Larger number of model components ($K\in\{30,50,80,100\}$) result in model spaces from millions to $\mathcal{O}(10^{30})$ configurations with up to $d_{\max} = 223$ parameters, where per-model inference and evidence-based BMC are infeasible. 
In these settings, we train for a fixed time frame of 24 hours and then evaluate (irrespective of convergence). Up to model spaces of $2^{30}$, the noise floor adjusted RMSE (rRMSE) is close to zero (Fig.~\ref{fig:symbolic_res}c). Performance drops afterwards, which is expected as only a small subset of the model space is seen during training, but remains good relative to the task complexity (e.g. prior rRMSE, see Appendix \ref{sec:appen_symbolic_additional_results} for a discussion). 
We evaluate posterior calibration via SBC \citep[Appendix \ref{app:subsec:eval_metrics}]{talts2018validating}. Predictions remain well-calibrated even for large model spaces (Fig.~\ref{fig:appendix_symbolic_metrics}). However, deviations from the Bayesian posterior measured via relative Kernel Stein Discrepancy (rKSD, \citet{liu2016ksd}, Appendix  \ref{app:subsec:eval_metrics})  increases at the largest scales. We note that increasing network capacity mitigates this effect and consistently improves performance in both data-rich and data-limited regimes (Fig.~\ref{fig:appendix_symbolic_metrics}). 
We further investigate model selection by constraining evaluation to a 200-model subspace and interpreting it as a 200-class classification problem in which data from many classes can be near-equivalent (i.e. redundant, functionally similar component combinations with a lot of noise). We therefore report top-5 accuracy, measuring whether the true model lies among the five highest-probability candidates. Top-5 accuracy remains high across scales 
($>90\%$, Tab.~\ref{tab:appendic_symbolic_classifications_metrics}), and we see a weak block structure in the confusion matrix (Fig.~\ref{fig:appendix_confussion matrix})   indicating that \ours\  reliably concentrates posterior mass on a small set of plausible structures even when the full combinatorial space is underexplored. This shows that \ours\ can effectively leverage shared structure across related models to \emph{generalize} to previously unseen configurations, yielding accurate and well-calibrated predictions even in under-sampled combinatorial spaces. 
\vspace{-0.1cm}

\subsection{Case study: fiber orientation reconstruction using diffusion MRI}
\label{sec:dmri_results}
{\setlength{\columnsep}{6pt} 
\begin{wrapfigure}{r}[0pt]{0.12\textwidth}
  \includegraphics[width=\linewidth]{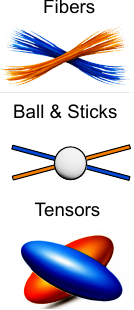}
  \vspace{-0.5cm}
  \caption{dMRI models.}
  \vspace{-0.1cm}
  \label{fig:illustrative-dmri}
\end{wrapfigure}

We demonstrate applications of \ours~ for biophysical model inference in \emph{in-vivo} diffusion MRI (dMRI) neuroimaging data. DMRI is a non-invasive imaging technique that provides a powerful method to map microstructure and connectional architecture of the brain. This requires solving an inverse problem; biophysical models are used to link the diffusion profile of water molecules within brain tissue measured by the MR scanner to the microstructural features of interest that hinder or restrict water diffusion. Over the past two decades,  dMRI has accumulated a broad ecosystem of such signal models (see \citet{panagiotaki2012compartment} for a review). Most models are built as combinations of primitive compartments (e.g., isotropic `ball' components to represent restricted diffusion within cell bodies,  anisotropic Gaussian compartments such as stick/tensor/zeppelin to define directionality, etc.). The task at hand, for example, is to select the number of `stick' compartments in each voxel of the image that represent different axonal fiber orientations given the data (Fig.~\ref{fig:illustrative-dmri}). 

The structure of this problem with hundreds of thousands of voxel-wise estimations per brain (i.e. voxel-wise fittings) is inherently suited for inference amortization. Each dMRI dataset comprises multiple acquisitions with given diffusion-encoding settings, conventionally summarized by a gradient direction
$\bm{\vec{b}}$ (`\texttt{bvec}') and a diffusion-weighting strength $b$ ( `\texttt{bval}') (Fig.~\ref{fig:eval_b3s}a). Recent simulation-based approaches have shown the potential of amortized frameworks for solving inverse problems in dMRI \cite{eggl2024more,eggl2025simulation,manzano2024uncertainty}. However, learning in these prior studies is restricted to a fixed model with a fixed acquisition protocol, often tied to a specific noise level and model, thus only providing inference over parameters (but not on multiple models). \ours~ can address all these limitations, allowing for rapid joint inference on models and model parameters that is amortized across voxels, dMRI acquisition settings, and noise settings. 
\ours\ is fully trained on simulations, but we also evaluate on real-world data, namely to UK Biobank (UKB) \cite{sudlow2015uk} and HCP \cite{van2012human} datasets.



\subsubsection{Ball-and-Sticks Model}

\begin{figure}[t]
    \centering
    \vspace{-0.0cm}
    \includegraphics[width=0.98\linewidth]{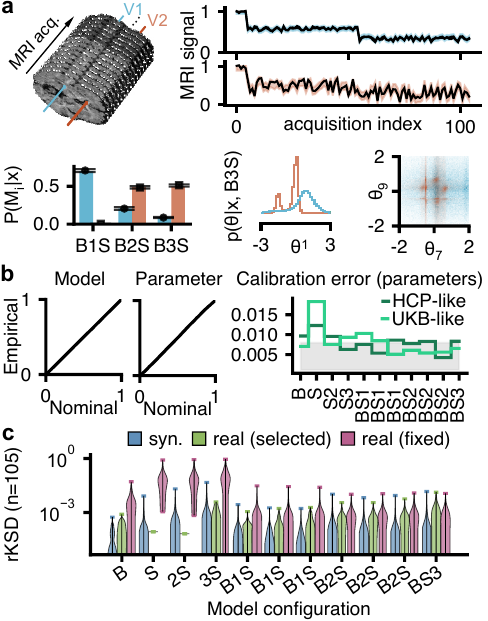}
    \caption{\textbf{\ours~ on dMRI. (a)} Two example voxels (V1, V2) with  acquired dMRI signal (black),  posterior predictives (color; 95\% quantile) and complex multimodal parameter posteriors (bottom, right, all marginals in Fig.~\ref{fig:b3s_pairplot}). Model posteriors (bottom, left) closely match unbiased Monte Carlo estimates (black). 
    \textbf{(b)} \emph{Left:} Simulation-based calibration over $10^{5}$ simulations averaged over all $\lambda$, $\vm$ and $\vtheta_\vm$.
    \emph{Right:} Calibration error for individual models across the model space for two acquisition schemes (gray: expected error), indicating excellent calibration.
    \textbf{(c)} rKSD distance to the posterior on synthetic and experimental data for B3S-family models, for a 'fixed' or a 'selected' model on UKB data (see Fig.~\ref{fig:app:dMRI_calibration_rKSD} for HCP data).
    (Models with same label refer to differences in prior.)
    }
    \label{fig:eval_b3s}
\end{figure}

Following the specifications of \citet{manzano2024uncertainty} (Appendix \ref{app:dMRI}), we consider the multi-shell Ball-and-Sticks model (B3S) \citep{behrens2007multifibre, jbabdi2012model}.
Here, the normalized attenuation of the dMRI signal $S/S_0$ is explained as a combination of an isotropic \emph{ball} compartment,
$g_{B}(b,\bm{\vec{b}})=\exp(-bd)$, 
and a weighted sum of up to N anisotropic \emph{sticks}:
$g_{S}(b,\bm{\vec{b}})=\exp\!\big(-bd\langle \bm{\vec{b}},\bm{\vec{\mu}}\rangle^{2}\big)$.
Model parameters include the diffusivity $d$, stick orientations $\bm{\vec{\mu}}$, and weights $f_k$ of each $k=1,...,K$ compartments.

The inference network is conditioned on diffusion-weighted measurements $\bm{S}=[s_1,\dots,s_n]$ (Fig.~\ref{fig:eval_b3s}a) and the corresponding acquisition settings $(b,\bm{\vec{b}})$, enabling amortization across acquisition protocols.
Since $\bm{S}$ is an exchangeable set indexed by acquisition settings (not an IID sequence), we encode it with a shallow (2-layer) transformer without position embeddings.
We use the same model prior as in Sec.~\ref{sec:symbolic_res}. We amortize over a broad range of randomly generated in-vivo-like protocols, varying b-values (up to $b=$6000~s/mm$^2$), gradient directions $\bm{\vec{b}}$, and number of acquisitions ($n$), enabling deployment across heterogeneous datasets. We consider models with $N=0,1,2$ and $3$ stick compartments, amortizing across a range of fiber complexity representations.

\textbf{Joint-posterior inference performance.}
For each voxel $\text{V}_i$, we infer a posterior over models and, conditional on the model probabilities, a posterior over its parameters in an amortized manner (Fig.~\ref{fig:eval_b3s}a).
The parameter posterior admits a complex multimodal geometry due to mixture symmetries and the posterior predictives closely match the observations (Fig.~\ref{fig:eval_b3s}a, right, Fig.~\ref{fig:b3s_pairplot},~\ref{fig:all_pairplot}).
The inferred model posterior aligns closely with likelihood-based estimators of the evidence (Fig.~\ref{fig:eval_b3s}a, down left; Fig.~\ref{fig:b3s_model_selection_eval}, Appendix Sec.~\ref{sec:app_dMRI_eval}), and both the model as well as the parameter posteriors are well calibrated across the full amortization scope in terms of SBC, i.e. it is neither over- nor underconfident when identifying models or parameters (Fig.~\ref{fig:eval_b3s}b). 

To complement SBC with a discrepancy measure applicable to real experimental data, we apply rKSD to the parameter posterior (Fig.~\ref{fig:eval_b3s}c).
rKSD confirms excellent approximation on synthetic data, as well as on experimental data within the full \ours~ pipeline. 
Importantly, if we restrict evaluation to a specific model class as a control experiment, it reveals larger deviations under overly simple (or inappropriate) models, consistent with expected performance degradations of SBI under model misspecification \citep{cannon2022investigating}.
This analysis shows the importance of model selection for trustworthy predictions in this task. 

Overall, these results show that the inferred model posterior closely matches exact evidence-based BMC ($R^2=0.97$; Fig.~\ref{fig:b3s_model_selection_eval}), and that the parameter posterior is near the true posterior. 
Moreover, high effective sample sizes ($\approx 60\%$ on synthetic data; $\approx 40\%$ on real data, Tab.~\ref{tab:appendix_b3s_performance}) suggest that remaining discrepancy can be corrected efficiently, e.g., via importance sampling \citep{Dax_2023}.

\begin{figure}[tp]
    \centering
    \vspace{-0.cm}
    \includegraphics[width=0.96\linewidth]{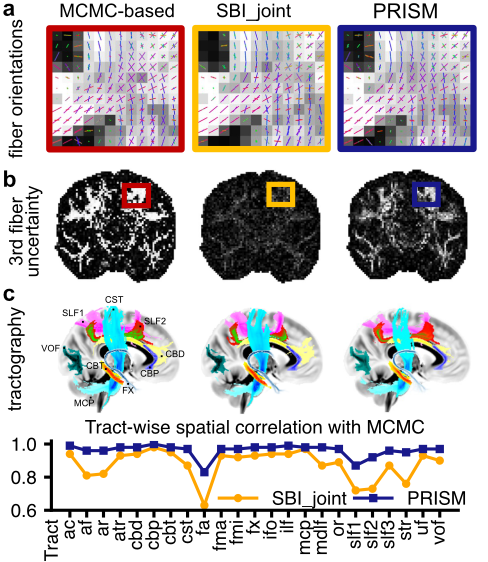}
    \vspace{-0.12cm}
    \caption{\textbf{Downstream Tractography. Comparisons against MCMC (BedpostX) and previous SBI approaches. (a)} 3-way fiber crossings from the Centrum Semiovale area. \textbf{(b)} Uncertainty of the third stick (fiber) orientation in terms of dispersion. High uncertainty (dark) indicates the third fibre is not supported by the data.
    \emph{Last row:} Tractography of brain regions based on inferred posteriors.
    \textbf{(c)} Tractwise correlation with MCMC, compared to SBI$_\text{joint}$ (tract labels from a). }
    \label{fig:eval_tract}
    \vspace{-0.65cm}
\end{figure}

\textbf{Probabilistic tractography.}
The posterior distributions of fiber orientations can be used to identify probabilistic white-matter paths of anatomical connectivity. We use XTRACT \cite{warrington2022concurrent} to evaluate probabilistic tractography. We compare our reconstructed bundled tracts against the ones using posterior estimates from BedpostX \citep{behrens2003characterization}, an extensively-validated MCMC approach used as the reference, and the SBI$_\text{joint}$ approach with restricted model choice proposed in \citep{manzano2024uncertainty}.

%
Compared to the SBI$_\text{joint}$ baseline, \ours~ improves agreement with MCMC and increases coherence in crossing-fiber regions (Fig.\ref{fig:eval_tract}a). Especially the uncertainty of fiber orientations is better matched and yields sharper contrast between white- and non-white matter (Fig.\ref{fig:eval_tract}b).
These improvements directly translate into higher spatial correlation of the reconstructed bundle tracts with MCMC across brain regions (Fig.~\ref{fig:eval_tract}c), with mean correlations of 0.96 versus 0.86 for the SBI baseline.  In addition, \ours~ can be applied to a much larger set of datasets (i.e. arbitrary $b$, $\bm{\vec{b}}$) without retraining. However, the computational cost for \ours~ generally increases compared to SBI$_\text{joint}$ (based on normalizing flows) and benefits from  GPU acceleration. 

Overall, generating one thousand model posterior samples takes 5~ms and 50~ms for the parameter posterior (on Nvidia H100, Fig.~\ref{fig:runtimes}), allowing inference on e.g. the UKB dataset within 4 minutes (in batches of 30k voxels).


 \subsubsection{Extended model classes}

Finally, to assess whether \ours\ can scale to larger and heterogeneous model spaces, we extend the approach to a substantially broader component library. We train a \emph{single} inference network amortized across the entire model space, which can either be queried for a specific model of interest, or to perform model selection over a model (sub)space. This setup is intended as a proof of principle demonstrating scalability and flexibility, rather than as a claim about optimal model choice.
We extend B3S with three Zeppelin/Tensor components forming the BSZT space (and further spherical-convolution components e.g.~\citet{zhang2012noddi,sotiropoulos2012ball} to obtain BSZT+conv, Sec.\ref{app:dMRI}, Tab.~\ref{tab:dmri_models_priors_params}). In total, this library then comprises 21 model components that can be grouped combinatorially, yielding a model family with up to 114 parameters (compared to up to 12 parameters in the B3S setting). Since candidates differ in parameterization, we penalize model complexity by parameter count via a dimension-penalizing prior (Sec.~\ref{sec:bszt_task_configuration}) 

We first verify on synthetic data that inference is well calibrated (Fig.~\ref{fig:eval_all}a, with minor deviation for BSZT+conv, Fig.~\ref{fig:appendix_all_calibration_error}) and achieve generally low predictive RMSE and KSD, but a drop in ESS (Tab.~\ref{tab:appendix_all_performance},~\ref{tab:appendix_all_conv_performance}). 
\ours\ distinguishes models when they are sufficiently different, but, as expected, it cannot reliably distinguish similar or equivalent models (Fig.~\ref{fig:confusion_matrix_all_all_conv}, Tab.~\ref{tab:appendic_dmri_classifications_metrics_merged}). We ran model selection on UKB data (Sec.~\ref{sec:app_model_selection}), and investigated whether some fixed models or the full Bayesian model average had strong statistical support on empirical data: Inferred models outperform DTI~\citep{basser1994dti} and maximum-likelihood methods such as Rumba~\citep{garyfallidis2014dipy,canales2015spherical} in signal reconstruction and leave-one-out cross-validation (Fig.~\ref{fig:eval_all}b). Mean performance is comparable among inferred models.  Yet they often induce different uncertainty in fiber orientation estimates (Fig.~\ref{fig:eval_all}c; Fig.~\ref{fig:appendix_fod}, Sec.~\ref{fig:appendix_fod}): principal directions largely agreed, but uncertainty could vary substantially. 

Overall, these results demonstrate that \ours\ can amortize joint model-parameter inference across large, heterogeneous, and combinatorial model spaces which would be intractable with per-model approaches. Can be used to discover data-consistent models within large model spaces (Sec.~\ref{sec:app_model_selection}, Fig.~\ref{fig:app:model_selection}) and, furthermore, can be applied to different experimental setups (e.g. acquisition protocols). Even in the setting of  noisy, real world data, it yields model- and parameter-identifications that remain well calibrated, have competitive statistical performance on real data, and yield interpretable uncertainty estimates.

\begin{figure}[t]
    \centering
     \vspace{-0.cm}
    \includegraphics[width=\linewidth]{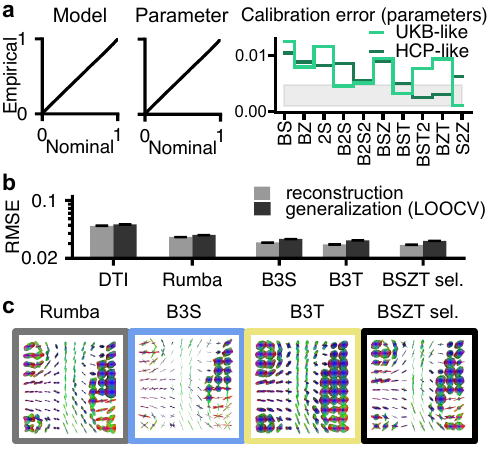}
    \vspace{-0.5cm}
    \caption{ \textbf{\ours~ on an extended model family for dMRI.}
    \textbf{(a)} Calibration error for the BSZT model on average (left) and across model classes for the parameter posterior (right).
    \textbf{(b)}  Reconstruction and generalization RMSE for reference models as well as for B3S, B3T,  BSZT sel. (i.e. the Bayesian model average) (generalization estimated by leave-one-out cross-validation, LOOCV).
    \textbf{(c)} Examples of predicted fiber orientation distributions (Sec.~\ref{sec:dmri_model_implemenation_details}) for a local patch.
    }
    \label{fig:eval_all}
    \vspace{-0.51cm}
\end{figure}

\section{Related Work}

In Approximate Bayesian Computation (ABC), likelihood-free model choice has been implemented by augmenting inference with a discrete model index and estimating posterior model probabilities via rejection sampling or SMC \citep{toni2009abcsmc,toni2010simulationbased,liepe2010abcsysbio}. However, this approach  requires sufficient summary statistics for consistency \citep{robert2011lackconfidence, marin2014relevant}.
ABC-RF further recasts model choice as supervised classification \citep{pudlo2016abcrf}. Neural SBI amortizes inference and often uses classifier-like model selection  \citep{pudlo2016abcrf} to infer posteriors over discrete model indices \citep{radev2021amortized}, with recent work addressing accuracy and validation for moderately sized candidate sets \citep{kucharsky2025improving,elsemueller2024deep,schumacher2025validation}, but omitting the parameter inference task.  The need to perform explicit enumeration during training/evaluation, however, limits scalability to combinatorially large spaces. Model comparison can also be performed via evidence estimation using NPE/NLE/NRE plug-in estimators \citep{spuriomancini2023bayesian} or flow-based normalizer recovery from unnormalized densities (e.g., $\tilde{p}/q$) \citep{srinivasan2024bayesian}. “All-in-one” systems learn per-model posteriors and (surrogate) likelihoods to enable evidence computation \citep{gloeckler2024allinonesimulationbasedinference, radev2023jana}, with COMPASS~\citep{gunes2025compassmodelcomparisonsimulationbased} extending this direction for model selection (but still requires per-model training and evaluation). Finally, SBMI \citep{schroder2024simultaneous} performs joint inference over model identity and parameters for large model collections, yet relies on restrictive approximation families and modeling choices (e.g., analytic marginalization of inactive dimensions). Mixture density networks can be limiting for complex posteriors and are often outperformed by more expressive generative models such as normalizing flows, diffusion models, and autoregressive decoders~\citep{papamakarios2021normalizing,karras2022elucidatingdesignspacediffusionbased}. Symbolic regression seeks interpretable equations from data. Early methods relied on genetic programming~\citep{schmidt2009distilling} or sparse regression over fixed bases~\citep{brunton2016discovering_SINDY,bakarji2023discovering_AE}. More recent deep-learning approaches use graph neural networks~\citep{cranmer2020discovering_graph_NN} or transformers~\citep{biggio2021neuralsymbolicregreassionthatscales}, but are typically deterministic and do not support systematic model comparison.

\vspace{-0.1cm}

\section{Discussion}

We introduced \ours, an amortized simulation-based inference framework for \emph{joint inference} over discrete model structure and continuous parameters in large, combinatorial model families. By learning a joint posterior $p(\mathcal{M},\theta\mid x,\lambda)$, \ours\ supports not only model selection but also \emph{model discovery} and efficient \emph{marginalization over model uncertainty} within a single amortized system. One advantage of amortization is the ability to perform rapid \emph{Bayesian model averaging} across large sets of plausible models. Rather than conditioning conclusions on a single model, predictions and downstream quantities can be marginalized over model identity, weighted by posterior probabilities. This makes uncertainty due to competing mechanistic hypotheses explicit and is particularly valuable in settings requiring repeated or time-critical inference.

\ours\ enables controlled exploration of model space via a tunable hierarchical prior
$p(\mathcal{M}\mid \lambda)$ that trades off complexity and fit. Importantly, this trade-off needs not be fixed at training time. Choosing $\lambda$ \emph{a priori} is often difficult,
and its effect on operational complexity (e.g., the number of active components) can be
highly nonlinear. This situation mirrors a well-known issue in nonparametric Bayesian methods \cite{escobar1995bayesian, teh2006hierarchical, ghahramani2013bayesian}, where hyperparameters implicitly controlling model complexity (e.g., number of clusters in nonparametric clustering) are frequently adjusted after inspecting inferred solutions. By conditioning on $\lambda$, \ours\ allows users to either fix it when strong prior knowledge is available, or to systematically explore its effect post hoc, e.g. using cross-validation.

These benefits come with limitations: Amortized joint inference requires substantial simulation budgets.  In large combinatorial model spaces only a small fraction of possible configurations is explored during training. While \ours\ generalizes well under such undersampling by leveraging shared structure across models, posterior quality can degrade when too large regions of model space remain unseen. In addition, large scale amortization benefits from larger neural networks with larger computational demands.

Despite these challenges, \ours\ shows that amortized inference can scale model discovery, selection, and averaging to regimes that are inaccessible to classical per-model Bayesian workflows, as well as across different measurement schemes, while retaining calibrated uncertainty over both model identity and parameters. \ours\ provides a practical foundation for simulation-based scientific inference in rich and heterogeneous model- and data-spaces, and which we expect to be useful across multiple scientific disciplines.

\section*{Software and Data}
We used \texttt{jax} \citep{jax2018github} as backbone, \texttt{hydra} ~\citep{Yadan2019Hydra} to track all configurations as well as several utilities from \texttt{sbi}~\cite{boelts2024sbi}. Code to reproduce results is available at \url{https://github.com/mackelab/prism}.

\section*{Acknowledgments}
We thank Stefan  Wahl for  feedback on the manuscript. We thank all members of the Mackelab for discussions and feedback on the manuscript. This work was funded by the German Research Foundation (DFG) under Germany’s Excellence Strategy – EXC number 2064/1 – 390727645 and SFB 1233 `Robust Vision' (276693517), the German Federal Ministry of Education (Tübingen AI Center) the European Union (ERC, DeepCoMechTome, 101089288), the ``Certification and Foundations of Safe Machine Learning Systems in Healthcare'' project funded by the Carl Zeiss Foundation. MG is member of the International Max Planck Research School for Intelligent Systems (IMPRS-IS). JP and SS are supported by an ERC Consolidator Grant (101000969), and JP is also supported by a Wellcome Trust, UK bioimaging technology award (313367/Z/24/Z).


\bibliography{references}
\bibliographystyle{icml2026}

\newpage
\appendix
\onecolumn
\renewcommand{\thefigure}{A.\arabic{figure}}
\setcounter{figure}{0}

\section{Experimental details}

We first define the metrics we used to assess approximation quality across the amortization scope (Sec.~\ref{app:subsec:eval_metrics}).
We then provide additional training details (Sec.~\ref{app:model_training_details}).

\subsection{Evaluation metrics}
\label{app:subsec:eval_metrics}

Having both a \textit{model} posterior and a \textit{parameter} posterior enables uncertainty quantification
at two levels: \emph{local} uncertainty conditioned on a fixed model, and \emph{global} uncertainty that
marginalizes over model uncertainty \citep{werner2021informed}.
Concretely, we evaluate the two posterior predictive distributions:
\begin{align}
\label{eq:global_local_predictive}
    \begin{split}
       \text{Local~: } \; & p(\vx \mid \vx_o,\vm)
       \;\approx\; \mathbb{E}_{q(\vtheta \mid \vx_o,\vm)}\!\left[p(\vx \mid \vtheta_\vm,\vm)\right],\\
       \text{Global: } \; & p(\vx \mid \vx_o)
       \;\approx\; \mathbb{E}_{q(\vm,\vtheta \mid \vx_o)}\!\left[p(\vx \mid \vtheta_\vm,\vm)\right].
   \end{split}
\end{align}

The model posterior further supports Bayesian model comparison (BMC) over any user-specified subset
$\{\vm_i\}\subset\vM$ by ranking models via $q(\vm_i \mid \vx_o,\lambda)$
\citep{radev2020bayesflow,elsemueller2024deep}. Finally, it enables \emph{model discovery} by exploring the
combinatorial space through samples $\mathcal{M}\sim q(\mathcal{M}\mid \vx_o,\lambda)$, surfacing plausible
component combinations beyond hand-designed candidates.

However, to measure performance we report the following quantities since exact reference posteriors are infeasible across our amortization scopes:
\begin{enumerate}[(i)]
    \item  \emph{Predictive RMSE}: We report \emph{local} RMSE under a fixed model and \emph{global} RMSE under the model-averaged posterior predictive  (\citealt{gelman1996ppc}). We report relative RMSE (rRMSE) to the lower bound (noise floor) where appropriate.
    \item  \emph{Calibration (SBC)}: Simulation-based calibration (SBC) assesses self-consistency of the inferred posterior by testing rank uniformity for parameters and models drawn from the generative process \citep{talts2018validating}. We report the absolute deviation of the SBC statistic from its nominal value. A well calibrated posterior is neither over- nor under-confident. However, calibration is a necessary, but not sufficient criterion for the correctness of the posterior distribution.
    \item  \emph{Sample quality via (r)KSD}: When likelihood scores are available, we evaluate posterior sample quality using (kernelized) Stein discrepancy, which does not require the posterior normalizing constant \citep{liu2016ksd,gorham2017measuring}. We report a normalized rKSD to mitigate scale effects across models and parameter dimensionalities .
\end{enumerate}

\paragraph{Predictive RMSE.}
For an observation $\vx_o$ and a fixed model $\vm$, we form the \emph{local} posterior predictive and for the full model family the \emph{global} (model-averaged) posterior predictive following Equation \ref{eq:global_local_predictive}.
We compute MSE on the predictive samples $\vx_i$ sampled from the local or global posterior predictive:
\begin{equation*}
\mathrm{MSE}(\vx_i,\vx_{o,i}) \;=\; \frac{1}{D}\sum_{j=1}^{D}\big(\hat{x}_i^j-x_{o,i}^j\big)^2,
\end{equation*}
where $D$ is the dimensionality of $\vx_o$. For \emph{local} MSE (i.e. conditioned on a model) we use $\vx_i
\sim\mathbb{E}_{q(\vtheta\mid \vx_o,\vm)}[p(\vx\mid\vtheta,\vm)]$, and for \emph{global} RMSE we use
$\vx_i \sim \mathbb{E}_{q(\vm,\vtheta\mid \vx_o)}[p(\vx\mid\vtheta,\vm)]$. For a dataset of $N$ observations we obtain the RMSE
$$ \text{RMSE} = \sqrt{\frac{1}{N} \sum_{i=1}^N \text{MSE}(\vx_i, \vx_{o,i})} $$

As this includes irreducible error from noise within the simulator we normalize RMSE by a task-specific irreducible-error baseline. Concretely, for each ground-truth setting
$(\vm^\star,\vtheta^\star)$ we generate $R$ replicate observations
$\vx_o^{(r)}\sim p(\vx\mid \vtheta^\star,\vm^\star)$ with different simulator seeds. Computing the RMSE on such a dataset yields $\mathrm{RMSE}_{\min}$. We report $\mathrm{rRMSE} = \mathrm{RMSE}-\mathrm{RMSE}_{\min}$ so values close to $0$ indicate performance near the simulation-noise floor.

\paragraph{Calibration (SBC).}
Simulation-based calibration (SBC) checks self-consistency under the generative process: if
$(\mathcal{M},\vtheta,\vx)$ are drawn from the simulator and we rerun inference on $\vx$, then suitable probability integral transform (PIT), e.g. rank statistics should be $\mathrm{Uniform}(0,1)$ \citep{talts2018validating}. We apply SBC to both the
\emph{model posterior} $q(\mathcal{M}\mid \vx)$ (discrete) and the \emph{parameter posterior}
$q(\vtheta\mid \mathcal{M},\vx)$ (continuous, conditional on the generating mask).
Notably this only allows performance evaluation of the  \textit{uncertainty calibration} in the \textit{well-specified} (i.e. synthetic) case.

More specifically, we sample for each trial $t=1,\dots,T$ model, parameters and data:
$\mathcal{M}^{(t)}\sim p(\mathcal{M})$, $\vtheta^{(t)}\sim p(\vtheta\mid\mathcal{M}^{(t)})$,
$\vx^{(t)}\sim p(\vx\mid\vtheta^{(t)},\mathcal{M}^{(t)})$, and then compute the probabilities
$q(\mathcal{M}\mid \vx^{(t)})$ and $q(\vtheta\mid \mathcal{M}^{(t)},\vx^{(t)})$.
We then compute SBC for the model masks and the parameters separately:

\emph{Model SBC (discrete).}
Let $\ell_{\text{true}}^{(t)}=\log q(\mathcal{M}^{(t)}\mid \vx^{(t)})$. Draw $S$ masks
$\widetilde{\mathcal{M}}^{(t,s)}\sim q(\mathcal{M}\mid \vx^{(t)})$ and set
$\ell_s^{(t)}=\log q(\widetilde{\mathcal{M}}^{(t,s)}\mid \vx^{(t)})$.
We use a randomized rank to handle ties: define
$n_{<}^{(t)}=\sum_s \mathbf{1}[\ell_s^{(t)}<\ell_{\text{true}}^{(t)}]$ and
$n_{=}^{(t)}=\sum_s \mathbf{1}[\ell_s^{(t)}=\ell_{\text{true}}^{(t)}]$,
draw $K^{(t)}\sim\mathrm{Unif}\{0,\dots,n_{=}^{(t)}\}$ and $V^{(t)}\sim\mathrm{Unif}(0,1)$, and form
$u^{(t)}_{\text{model}}=(n_{<}^{(t)}+K^{(t)}+V^{(t)})/(S+1)$. This is important in discrete cases as there is a non-zero probability to sample the exact same mask with exact same log probabilities.

\emph{Parameter SBC (continuous, conditional on the generating model).}
Let $\mathcal{I}(\mathcal{M}^{(t)})$ be the active parameter indices under $\mathcal{M}^{(t)}$, and draw
posterior samples $\vtheta^{(t,s)}\sim q(\vtheta\mid \mathcal{M}^{(t)},\vx^{(t)})$.
For each $j\in\mathcal{I}(\mathcal{M}^{(t)})$, compute the rank of the truth
among samples,
e.g.\ $r^{(t)}_j = 1+\sum_s \mathbf{1}[\theta^{(t,s)}_j < \theta^{(t)}_j]$, and convert to a PIT value
$u^{(t)}_{j}=(r^{(t)}_j-0.5)/(S+1)$. We then assess uniformity by pooling all
$\{u^{(t)}_{j}\}$ across trials and active indices .

\emph{Calibration summary.}
Given PIT values $\{u\}$ (either $\{u^{(t)}_{\text{model}}\}$ or the pooled $\{u^{(t)}_{j}\}$), we compare the
empirical CDF $\widehat{F}(u)$ to the nominal $F_0(u)=u$. As a scalar error we report
$\mathrm{CE}=|\mathcal{G}|^{-1}\sum_{g\in\mathcal{G}}|\widehat{F}(g)-g|$ on a grid $\mathcal{G}\subset[0,1]$. We use a uniform grid with 100 vertices as default.

\paragraph{(Relative) multi-scale KSD.}
To quantify how well samples from the parameter posterior match the target posterior, we use a preconditioned, multi-scale
Kernel Stein Discrepancy (KSD, ~\citet{liu2016ksd, gorham2017measuring}) computed \emph{conditional on the generating model} $\mathcal{M}$.
For each trial (or voxel) we draw samples $\{\vtheta^{(s)}\}_{s=1}^S \sim q(\vtheta\mid \mathcal{M},\vx)$
and evaluate them against the unnormalized log posterior
$\log p(\vtheta\mid \mathcal{M},\vx)=\log p(\vx\mid\vtheta,\mathcal{M})+\log p(\vtheta\mid\mathcal{M})$,
where the prior is applied only to the active dimensions indicated by $\theta$-mask
$\mathbf{m}_\theta(\mathcal{M})$ (effective dimension $d_{\text{eff}}=\sum_j \mathbf{m}_{\theta,j}$).
We obtain the score $\nabla_{\vtheta}\log p(\vtheta\mid \mathcal{M},\vx)$ by automatic differentiation.

\emph{Preconditioning.}
To reduce sensitivity to anisotropy and scale, we whiten samples in the active subspace using an empirical
covariance preconditioner: with $L L^\top \approx \widehat{\mathrm{Cov}}(\vtheta\odot \mathbf{m}_\theta)$,
we set $P=L^{-1}$, transform samples to $ \vz = P\,\vtheta$, and transform scores accordingly
$\vs_{\vz}=P^{-\top}\vs_{\vtheta}$ (again masking inactive dimensions).

\emph{Multi-scale Stein kernel.}
In whitened space we use an RBF base kernel with multiple bandwidths
$k_h(\vz,\vz')=\exp(-\|\vz-\vz'\|^2/(2h^2))$ for $h\in\mathcal{H}$, and average the corresponding Stein
kernels uniformly across scales. We use  a broad range ($h\in \{0.1,0.5,1.,10.\}$) to increase sensitivity to a broader range of ``discrepancies''.  The (squared) KSD is computed as the usual
U-statistic over the resulting Stein Gram matrix $H$:
$$
\mathrm{KSD}^2 = \frac{1}{S(S-1)}\sum_{i\neq j} H_{ij}, \qquad \mathrm{KSD}=\sqrt{\max(\mathrm{KSD}^2,0)} .
$$

\emph{Relative KSD (rKSD).}
Because KSD magnitude depends on dimension and problem, we also report a standardized discrepancy.
Let $\mathrm{KSD}_{\text{prior}}$ be the KSD obtained by replacing posterior samples with prior samples
$\vtheta^{(s)}\sim p(\vtheta\mid\mathcal{M})$ (using the same $S$, mask, preconditioning, and bandwidth set).
We then define
$\mathrm{rKSD}=\mathrm{KSD}/\mathrm{KSD}_{\text{prior}}$,
so $\mathrm{rKSD}<1$ indicates the approximation is closer to the target than the prior baseline.

\subsection{Model architecture and training details}
\label{app:model_training_details}

\subsubsection{Model details and configurations}
The full model comprises (i) an \textit{encoder} for the simulator output and global parameters (e.g. diffusion measurements and acquisition meta-data), (ii) a \textit{tokenizer} that produces model component- and parameter-level token sequences together with structured attention masks, and (iii) two transformer \textit{decoders}: an autoregressive (AR) decoder for the model posterior $q_\phi(\mathcal{M}\mid \mathbf{x},\lambda)$ and a diffusion decoder for the parameter posterior $q_\phi(\boldsymbol{\theta}\mid \mathcal{M},\mathbf{x},\lambda)$.

We only vary the \texttt{model\_dim} and \texttt{num\_layer} across experiments, all other hyperparameters were fixed or derived from that. We used 4 attention heads, attention size is fixed to 16 (query/key/value projection dimension), and widening factor 4 (i.e. feedforward MLP expand the hidden dimension by 4 $\times$ \texttt{model\_dim}).

\textbf{Encoder.} Encoders across tasks are represented by a standard self-attention transformer encoder. For the symbolic regression task we group 10 datapoints into a single token and additionally add a positional embedding. For the dMRI task  we embed the acquisition settings i.e. $(b,\vec{\mathbf{b}})$ and diffusion signal $\bm{S}$ as follows: we use a Random Fourier Embedding for the b-value $b$, a linear mapping for the signal $\bm{S}$ and the bvec $\vec{\mathbf{b}}$ which all get concatenated into a token of size \texttt{model\_dim}, which is then transformed by a transformer without positional embedding (to ensure exchangeability).

\textbf{Tokenizers.} The learnable tokenizers have a similar structure but task-specific variations. All used tokenizers have a learnable ``identification'' vector per component in which the necessary information is provided for the decoders.

The model decoder additionally requires the binary mask $\mathcal{M}=(M_1,\dots,M_C)$; we encode this by adding a learned embedding of the mask values $M_i\in\{0,1\}$ to the corresponding token.

\begin{itemize}
    \item[(i)] \texttt{SymbolicTokenizer:} We have several parameters per model component (Tab.\ref{tab:extended_symbolic_components}). Each parameter of $\vtheta_{\vm} \subset \vTheta$  is linearly projected to a \texttt{model\_dim}-dimensional vector and  added to the component embedding. In addition to this embedding layer, we also need a component specific linear decoding layer to reduce the token to the respective parameter dimension.
    \item[(ii)] \texttt{DMRITokenizer} extends this to also allow for \textit{global} or \textit{shared} parameter per model component. For global parameters (i.e. shared diffusivity in B3S or the constraints on the fractions $\vf$) we introduce a learnable identification token per dimension and add a linear projection of the value (e.g. $f_i$) to each token. In a similar style \textit{shared parameters} get grouped into an additional token and are removed from the corresponding compartment tokens.
\end{itemize}

We additionally construct a $\mathcal{M}$-dependent \emph{block attention mask}: if $M_i=0$, the corresponding parameter-token block is removed from the diffusion-transformer input and all attention connections to and from that block are masked. This implements attention-based marginalization of unused parameters and ensures $\boldsymbol{\theta}$ predictions only depend on active components~\citep{gloeckler2024allinonesimulationbasedinference}.

\textbf{Model-posterior decoder.}
Discrete model selection is implemented with an \textit{autoregressive} transformer decoder, representing the model posterior as a multivariate Bernoulli distribution over $\mathcal{M}\in\{0,1\}^{C}$:
\begin{equation}
q_\phi(\mathcal{M}\mid \mathbf{x},\lambda)
=\prod_{i=1}^{C} \mathrm{Bern}\!\Big(M_i \,\Big|\, p_\phi\!\left(M_{<i}, \mathbf{x},\lambda\right)\Big).
\label{eq:ar_model_posterior_impl}
\end{equation}
Conditioning on $\mathbf{x}$ is implemented via cross-attention to the encoder tokens; $\lambda$ is injected through adaptive layer normalization (AdaLN) \citep{peebles2023scalable}. Specifically we embed $\lambda$ by a Random Fourier Embedding. Each AdaLN block then uses a MLP to predict a "scale and shift" of size \texttt{model\_dim} that then scales or shifts the transformers hidden state. In contrast to \citet{peebles2023scalable} we do not use additional gating. A causal attention mask (with the padding tokens) enforces strict autoregressive decoding. We can thus evaluate the probability within a single forward pass, but require $C$ forward passes for sampling it autoregressively. The evaluation hence is $\mathcal{O}(C^2)$ and naive sampling cost is $\mathcal{O}(C \cdot C^2)$ ($\mathcal{O}(C^2)$  when using KV caching at cost of memory).

\textbf{Parameter-posterior decoder.}
Continuous parameter inference uses an EDM-style \textit{diffusion} transformer  \citep{karras2022elucidatingdesignspacediffusionbased}. For a global parameter vector $\boldsymbol{\Theta}\in\mathbb{R}^{d_{\max}}$, we sample $t\in[10^{-4},80]$ and perturb parameters as
\begin{equation}
\boldsymbol{\theta}_t=\boldsymbol{\theta}+t\,\boldsymbol{\epsilon},
\qquad
\boldsymbol{\epsilon}\sim\mathcal{N}(\mathbf{0},\mathbf{I}),
\label{eq:edm_forward_impl}
\end{equation}
conditioning on encoder tokens via cross-attention and injecting $t$ via Gaussian--Fourier time embeddings and AdaLN. We train with $v$-prediction,
\begin{equation*}
\mathbf{v}_t = \alpha_t \boldsymbol{\epsilon} - \beta_t \vtheta,
\quad
\mathbf{v}_\phi(\vtheta_t,t,\vx)=\alpha_t\,\hat{\boldsymbol{\epsilon}}_{\phi,\vtheta_t,t,\vx}-\beta_t\,\hat{\vtheta}_{\phi,\vtheta_t,t,\vx}.
\end{equation*}
with $\alpha_t,\beta_t$ as the normalized SNR $\alpha_t = \frac{1}{\sqrt{1 + t^2}}, \beta_t = \frac{t}{\sqrt{1+t^2}}$. Note, that $\mathbf{v}_\phi(\vtheta_t,t,\vx)$ does not need two evaluations for $\hat{\boldsymbol{\epsilon}}, \hat{\vtheta}$. From $\hat{\vtheta}$ as given by \citet{karras2022elucidatingdesignspacediffusionbased} we obtain $\hat{\boldsymbol{\epsilon}} = (\vtheta_t - \hat{\vtheta}) / t $.

Sampling uses a 64-step EDM schedule on the probability flow ODE~\citep{song2020generative}. If not specified otherwise, we use an ODE based solver e.g. the exponential Adams–Bashforth solver \citep{Lu_2025}, which only requires a single forward pass per step. Given that we group parameters compartmentwise the number of tokens is thus $\mathcal{O}(C)$ and sampling cost $\mathcal{O}(T\cdot C^2)$, where $T$ is the number of sampling steps in the diffusion model.

To impose a consistent prior on $\vtheta$ across models, we \emph{reparameterize} each task-specific prior into a standard-normal latent space via a bijection. Concretely, we introduce latent variables $\vtheta \sim \mathcal{N}(\vzero,\mathbf{I})$ and define task parameters as $\vtheta_{\text{natural}} = T(\vtheta)$, where $T$ maps unconstrained Euclidean latents to the appropriate constrained domain (see below, e.g. intervals, manifolds). This ensures that samples always satisfy the task constraints (e.g., positivity, simplex constraints, or manifold membership), while allowing us to use a consistent normal prior across dimensions. Additionally this allows to reduce dimensionality to the domains dimension and reduces degeneracy.

We use the following bijective transforms within our experiments:
\begin{itemize}
    \item \textit{Uniform on an interval} $[a,b]$ (componentwise): for $\vtheta \sim\mathcal{N}(0,1)$, set $u=\Phi(z)\in(0,1)$ and
    $$
        \theta = a + (b-a)\,u ,
    $$
    where $\Phi$ is the standard-normal CDF.
    \item \textit{Uniform on the upper half-sphere:} $\mathbb{S}^2_{+}=\{\vn\in\mathbb{R}^3:\|\vn\|_2=1,\ n_z\ge 0\}$: draw $z_1,z_2\sim\mathcal{N}(0,1)$, set $u_1=\Phi(z_1)$, $u_2=\Phi(z_2)$, and define
    $$
        \varphi = 2\pi u_1, \qquad \theta = \arccos(u_2),
    $$
    $$
        \vn = \bigl(\sin\theta\cos\varphi,\ \sin\theta\sin\varphi,\ \cos\theta\bigr),
    $$
    which yields a uniform distribution over surface area on the hemisphere.
    \item \textit{Simplex} $\Delta^{K-1}$ (Dirichlet stick-breaking): for a Dirichlet prior $\bm{\pi}\sim\mathrm{Dir}(\bm{\alpha})$, we use the stick-breaking construction with $K\!-\!1$ latent variables $\bm{\epsilon}\in\mathbb{R}^{K-1}$.
    We map $\bm{\epsilon}$ to uniforms $u_i=\Phi(\bm{\epsilon}_i)$ and then to Beta variables via the inverse regularized incomplete beta function:
    $$
        v_i = I^{-1}_{u_i}\!\left(\alpha_i,\ b_i\right),\qquad
        b_i = \sum_{j>i}\alpha_j,
    $$
    followed by
    $$
        \pi_i = v_i \prod_{j<i}(1-v_j),\quad i=1,\dots,K-1,\qquad
        \pi_K = \prod_{j=1}^{K-1}(1-v_j).
    $$
\end{itemize}

\subsubsection{Online training}
Data is generated on the fly via a streaming \texttt{SimulationDataset}. A  simulator runs on CPU (if not otherwise specified) and fills a ring buffer of size $10^{5}$ in batch-size-sample chunks; a background thread refreshes consumed slots and host-to-device prefetch keeps the accelerator saturated. Training batches are fetched in a non-blocking manner indefinitely. Optimization uses RAdam (learning rate $5\cdot 10^{-4}$), adaptive gradient clipping at 2.0, no scheduler, and an exponential moving average (EMA) tracking of parameters (decay 0.999). The objective combines an auto-regressive mask negative log-likelihood (implemented as binary-cross-entropy) and a diffusion $v$-prediction loss:
\begin{equation}
\mathcal{L}_{\mathcal{M}}
=
-\mathbb{E}\Bigg[\sum_{i=1}^{C}\log \mathrm{Bern}\!\Big(M_i \,\Big|\, p_\phi(M_{<i},\mathbf{x},\lambda)\Big)\Bigg],
\qquad
\mathcal{L}_{\boldsymbol{\theta}}
=
\mathbb{E}_{t,\boldsymbol{\epsilon}}\Big[\big\|\hat{\mathbf{v}}_\phi(\boldsymbol{\theta}_t,t,\mathbf{x},\mathcal{M})-\mathbf{v}_t\big\|_2^2\Big].
\label{eq:training_losses_impl}
\end{equation}
Runs are executed on a single H100 GPU with 16 CPU cores, 32~GB host RAM. As final parameters we use the latest EMA parameter.

\subsubsection{Additional specifications}

For the symbolic regression tasks, we varied \texttt{model\_dim} $\in \{32,64,128\}$ while keeping the number of layers fixed. We used 2 layers for the encoder and 6 layers for the decoder, and trained with a batch size of 4096. For the comparison with SBMI, we used \texttt{model\_dim}$=64$, reduced the decoder to 4 layers, and reduced the batch size to 1024.

For the dMRI networks, we used \texttt{model\_dim}$=128$. The \textbf{B3S model} used layers $(2,4,6)$ for (encoder, model-selection, parameter inference), respectively, and the B3S simulator trained on a mixed acquisition scheme combining UKB and HCP-like data (Sec.~\ref{app:dMRI} for details). The \textbf{BSZT model} used layers $(2,6,8)$ for (encoder, model-selection, parameter inference), respectively, and trained on the same mixed acquisition scheme. The \textbf{BSZT+conv model} used \texttt{model\_dim}$=128$ and the same stack depths as the BSZT model, but with the spherical-convolution extension enabled in the simulator and training data generation. As the spherical convolution is much more expensive, we ran the simulation here also on GPU. All dMRI networks were trained for 72 hours.

\subsection{Additional symbolic regression results}
\label{sec:appen_symbolic_additional_results}

\begin{wrapfigure}{}{0.4\linewidth}
    \centering
    \includegraphics[width=\linewidth]{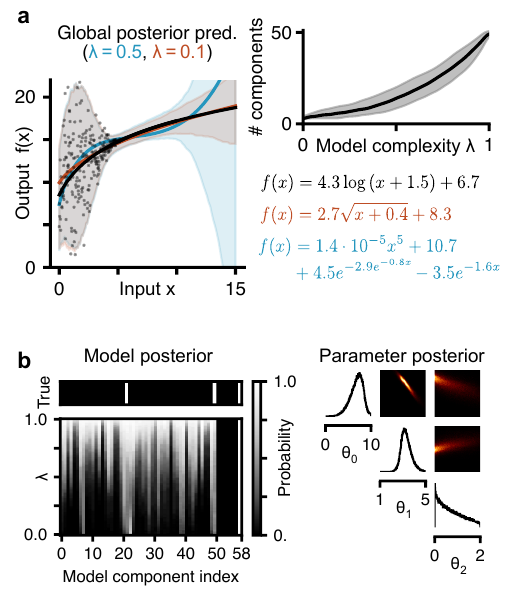}
    \caption{\textbf{(a)} Illustration of symbolic regression given on the same ground truth function and noise model as in Fig.~\ref{fig:symbolic_illustrative} but with 1000 observations.
    \textbf{(b)} Associated model and parameter posterior of the simple function (orange), which becomes more constrained compared to Fig.~\ref{fig:symbolic_illustrative} with 10 observations.}
    \label{fig:app:symbolic_illustrative_appendix}
\end{wrapfigure}

For the same example function as in Fig.~\ref{fig:symbolic_illustrative}, we show in Fig.~\ref{fig:app:symbolic_illustrative_appendix} the results for 1000 observations as also used in our main evaluations. We can see that the model posterior is more certain about the number of components (Fig.~\ref{fig:app:symbolic_illustrative_appendix}{a} right, Fig.~\ref{fig:app:symbolic_illustrative_appendix}{b} left) . Yet, due to large and 'peaked' noise for small $x$, the overall function is not identifiable and e.g. a square-root approximation to the log is similarly well supported by the data as the true logarithmic component.

For each trained model with $K \in \{15,30,50,80,100\}$ symbolic components, we evaluated rRMSE, rKSD, and calibration error (Fig.~\ref{fig:appendix_symbolic_metrics}). Both rRMSE and rKSD indicate a very good approximation up to $2^{30}$, with only a minor deviation at $2^{50}$ and a clearer deviation afterwards. Across all $K$, calibration error remains very low and is typically close to the error expected at random. For $K=15$ we observe larger (but still small) calibration deviations even though rRMSE and rKSD stay low. All in all, this does not imply that models with $K>50$ are inherently poor: they still yield accurate and calibrated predictions, but they deviate from the Bayes-optimal predictor.

\paragraph{Limited Training Data}
For the task of  $K=50$ model components, the smallest network completed 553k update steps, whereas the largest network completed roughly 363k. If we assume that each batch corresponds to a new model (i.e., no model is repeated; an upper bound), this amounts to approximately 1.4--2.2 billion simulations used for training. Even under the most generous assumption that each batch contains a distinct model configuration, the smallest network would have explored only about \(10^{-4}\%\) of the full model space, with essentially one data point per visited model. Consequently, good performance necessarily relies on \textit{extrapolation} across related models---this is the core benefit of \textit{large-scale amortization}. At the same time, the results suggest that achieving near Bayes-optimal performance requires covering at least a non-trivial fraction of the model space (e.g., for \(K=30\) we observed roughly two simulations per model configuration).

\begin{figure}[tp]
    \centering
    \includegraphics[width=\linewidth]{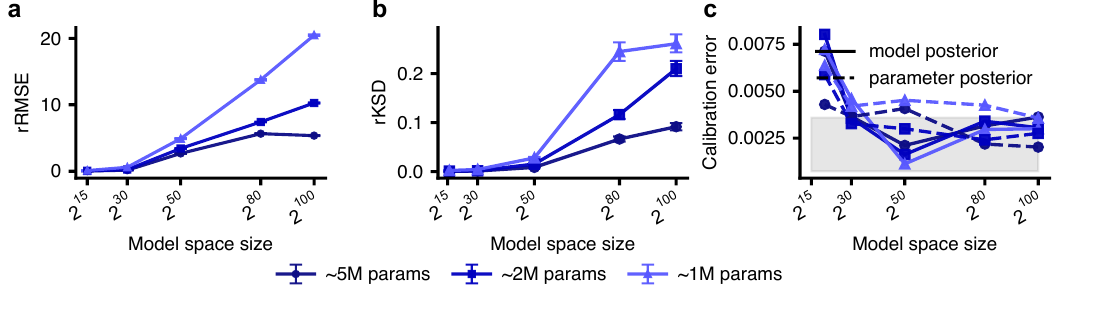}
    \caption{\textbf{Evaluation metrics across symbolic regression tasks.} The rRMSE (\textbf{a}), rKSD (\textbf{b}) and calibration error (\textbf{c}) across all considered model spaces (x-axis) and model sizes (colors). The gray shaded area indicates the expected Monte Carlo error if perfectly calibrated. Model sizes refer to \texttt{model\_dim} $\in\{32,64,128\}$.
    }
    \label{fig:appendix_symbolic_metrics}
\end{figure}

\paragraph{Classifier Perspective}
To assess the model-posterior network from a classification perspective, we treat each candidate model as a ``class.'' Because the full model space is too large for standard multi-class evaluation, we restrict evaluation to a random subspace \(\mathcal{M}_{200}\). Specifically, we sample 200 models (represented by their binary masks) and define the classifier
$$
c(\vm\mid \vx) = \arg\max_{\vm\in \mathcal{M}_{200}} q(\vm\mid \vx),
\qquad \text{with } \vx\sim p(\vx\mid \vm).
$$
We then evaluate this classifier using standard metrics. First, we compute confusion matrices for $K=30,50,100$ (Fig.~\ref{fig:appendix_confussion matrix}). We order models by similarity of their representative masks---i.e., by overlap in components rather than by direct functional equivalence. The resulting matrices are largely diagonal, indicating that the classifier typically recovers the true model, or a closely related one.

From these confusion matrices we derive standard classification metrics (Tab.~\ref{tab:appendic_symbolic_classifications_metrics}). Top-1 accuracy decreases monotonically with increasing $K$, from $0.796$ at $K=15$ to $0.503$ at $K=100$, with corresponding declines in macro precision, recall, and F1. Yet, low accuracy is expected: as $K$ grows, we have larger estimation error but also the number of plausible model configurations increases and many configurations become observationally near-indistinguishable, rendering exact identification of a single ``true'' model class increasingly ambiguous.

In this setting, Top-5 accuracy is a more informative measure of posterior concentration than Top-1 accuracy (as, e.g., also employed for ImageNet~\citep{russakovsky2015imagenetlargescalevisual}. For $K\in\{15,30,50\}$, the true model is almost always contained among the five highest-probability candidates (Top-5 accuracy \(\ge 0.981\)), indicating that $q(\vm\mid \vx)$ concentrates its mass on a very small set of alternatives even when the MAP choice is not always correct. For larger $K$, Top-5 accuracy remains high but declines (to $0.940$ at $K=80$ and $0.905$ at $K=100$). Importantly, this performance is achieved even though the training procedure cannot have covered all classes in $\vM$ (and unlikely all the classes in \(\mathcal{M}_{200}\)), so the classifier must generalize to previously unseen model configurations. Overall, these results suggest that the model-posterior network retains substantial discriminative power at large \(K\): while exact Top-1 accuracy is hard (or generally not possible) due to class degeneracy, the true model typically remains among a small set of highly plausible candidates.

\begin{figure}
    \centering
    \includegraphics[width=\linewidth]{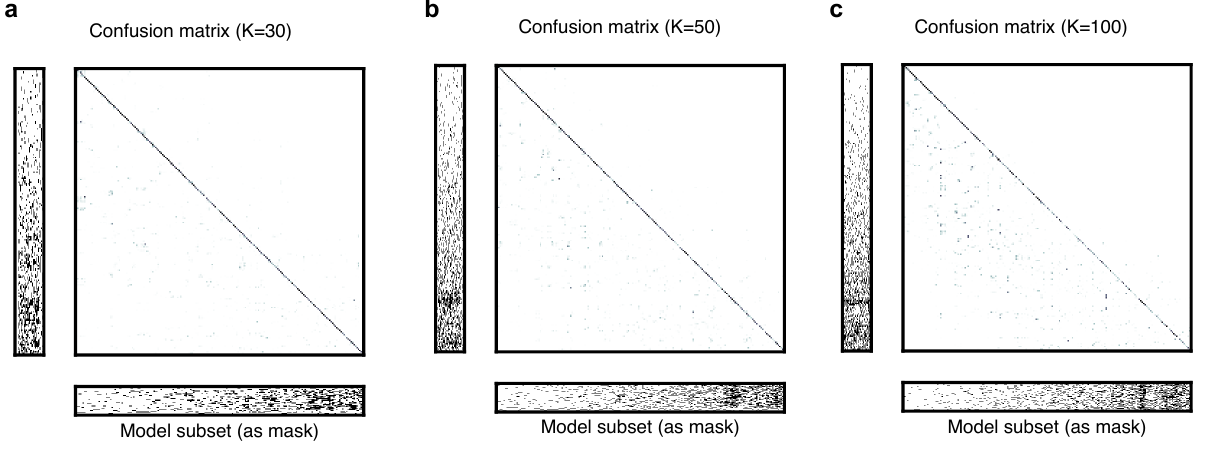}
    \caption{\textbf{Confusion matrix for model posterior network.} For a subset of 200 randomly selected models (indicated by their binary mask, x/y axis, grouped by similarity) we show the confusion matrix averaged over random $\lambda$ for the symbolic regression tasks with $K=30$, $K=50$, $K=100$}
    \label{fig:appendix_confussion matrix}
\end{figure}

\begin{table}[t]
\centering
\caption{Classification metrics for symbolic classification across different settings as derived from the confusion matrices in Fig.~\ref{fig:appendix_confussion matrix}. Please note that even a Bayes optimal classifier will spread mass across multiple models. }
\begin{tabular}{lccccc}
\hline
\textbf{Metric} & 15 & 30 & 50 & 80 & 100 \\
\hline
Accuracy (Top-1)  & 0.796 & 0.730 & 0.690 & 0.560 & 0.503 \\
Accuracy (Top-5)  & 0.981 & 0.995 & 0.995 & 0.940 & 0.905 \\
Macro precision   & 0.813 & 0.776 & 0.741 & 0.647 & 0.597 \\
Macro recall      & 0.797 & 0.742 & 0.688 & 0.548 & 0.503 \\
Macro F1          & 0.794 & 0.733 & 0.674 & 0.525 & 0.463 \\
\hline
\end{tabular}

\label{tab:appendic_symbolic_classifications_metrics}
\end{table}

\subsection{Additional dMRI results}
\subsubsection{Extended B3S inference performance evaluation}
\label{sec:app_dMRI_eval}

We further verify inference performance by evaluating parameter inference on both synthetic datasets and a real dataset (Tab.~\ref{tab:appendix_b3s_performance}). Notably, while our likelihoods are comparatively ``simple'', the induced posterior can still be highly complex---even within this constrained model family (Fig.~\ref{fig:b3s_pairplot}). This contrasts with other applications in simulation-based inference, where the likelihood (i.e.\ the simulator) is complicated but the associated posterior is often relatively simple~\citep{dax2021gravitational}.

In addition to KSD and RMSE, we report the effective sample size (ESS), which provides a proxy for how well approximate samples could be corrected toward the true posterior using, for example, importance sampling (cf.~\citep{dax2021gravitational}). We compute the ESS as follows: Let \(w_i = p(\vtheta_i \mid \vx)/q(\vtheta_i \mid \vx)\) denote importance weights for samples \(\vtheta_i \sim q(\vtheta \mid \vx)\). We use the normalized ESS,
\begin{equation}
\mathrm{ESS}
\;=\;
\frac{1}{N}\,
\frac{\left(\sum_{i=1}^N w_i\right)^2}{\sum_{i=1}^N w_i^2},
\end{equation}
where \(N\) is the number of samples. On synthetic data, we obtain normalized ESS values of approximately 60 \% on well-specified simulated data (Tab.~\ref{tab:appendix_b3s_performance}).
The ESS varies with diffusion steps of the sampling routine, but it quickly reaches its plateau with $T=32$ network evaluations which takes around 100 ms for 1000 posterior samples (Fig.~\ref{fig:runtimes}).

\begin{wrapfigure}{r}{0.35\textwidth}
        \centering
        \includegraphics[width=\linewidth]{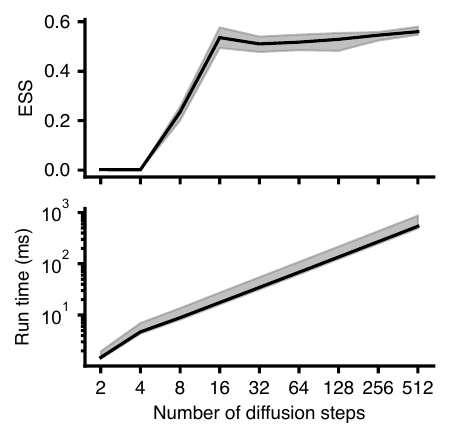}
        \vspace{-0.2cm}
        \caption{ESS and run times as a function of the number of diffusion steps for 1000 posterior samples for V2 B3S posterior in Fig.~\ref{fig:eval_b3s}. Evaluated on an Nvidia H100.}
        \label{fig:runtimes}
\end{wrapfigure}

We further evaluate the model posterior network under a stricter model-selection perspective. We therefore analyze the confusion matrix (Fig.~\ref{fig:b3s_model_selection_eval}{a}). While it is straightforward to separate a Ball from a Stick, distinguishing among Stick-based models is substantially harder when they induce (near-) identical data distributions. This ambiguity also propagates to closely related composite models (e.g.\ BS vs.\ S). We attribute this primarily to three factors:
First, we employ an informative fraction prior $\vf$; \emph{a priori}, the fraction associated with the third stick is already close to zero, so ``removing'' that component has only a minor effect on \(p(\vx \mid \vm)\). Second, we use a single noise model across a wide range of noise scales (or SNRs). If B2S and B3S are already similar in the noiseless setting, they become virtually indistinguishable at most noise levels, except for very small noise. Third, there are multiple B2S and B1S variants that are not exactly but nearly equivalent in our configuration (differing primarily in the fraction prior). Consequently, the B2S variants must split probability mass among themselves, whereas the B3S model does not face the same internal ambiguity.

To further validate these findings on real data, we compare the approximate model posterior against a ground-truth reference computed via evidence estimates on a random subset of 1000 voxels from UKB (e.g. dMRI data using the UK Biobank acquisition protocol). We estimate the evidence \(p(\vx_o \mid \vm)\) using an importance sampling estimator based on \(q(\vtheta \mid \vm, \vx_o)\), which yields low-variance estimates in this setting. Concretely, we compute
\begin{equation}
p(\vx_o \mid \vm)
\;\approx\;
\frac{1}{N}\sum_{i=1}^N
p(\vx_o \mid \vm, \vtheta_{\vm}^i)\,
\frac{p(\vtheta_\vm \mid \vm)}{q(\vtheta_\vm^i \mid \vm, \vx_o)},
\qquad
\vtheta_{\vm}^i \sim q(\vtheta_\vm^i \mid \vm, \vx_o),
\end{equation}
which is an unbiased Monte Carlo estimator of the evidence. Evidence estimates using the prior showed high variance, but this estimator with $N=1024$ samples showed a low enough variance for two independent estimators to closely agree ($R^2 = 0.99$, Fig.~\ref{fig:b3s_model_selection_eval}{b}).

As a complementary check, we also consider a biased evidence proxy derived from the model posterior network,
\begin{equation}
p(\vx_o \mid \vm)
\;\propto\;
\frac{q_\phi(\vm \mid \vx_o, \lambda)}{p(\vm \mid \lambda)},
\end{equation}
which is likewise well aligned with the unbiased importance-sampling estimates (\(R^2 = 0.97\), Fig.~\ref{fig:b3s_model_selection_eval}\textbf{c}). This demonstrates that models selected via the model-posterior network are well aligned with models selected through classical evidence-based model selection.

Given these evidence estimates, we can assess how well the approximate model posterior probabilities match the corresponding ground-truth model posterior,
\begin{equation}
p(\vm \mid \vx_o)
\;\propto\;
p(\vx_o \mid \vm)\,p(\vm \mid \lambda),
\end{equation}
across all \(\lambda\). We observe that the total variation distance (equivalently, the \(\ell_1\) distance up to a factor \(1/2\)) remains low overall, but increases for small \(\lambda\). This is consistent with model misspecification: for small \(\lambda\), the network is biased toward ``simple'' observations, whereas our earlier results suggest that real data is more ``complex'' in the sense that more complex models have higher evidence.

In conclusion, these experiments indicate that our model posterior approximation is accurate, and that the remaining unidentifiability is largely driven by the task configuration (Sec.~\ref{sec:b3s_task_configuration}). A straightforward modification to mitigate this issue is to replace the single noise model with multiple noise regimes (e.g.\ small, medium, and large noise), since identifiability improves at lower noise levels where model-induced distributions are more separable. This is consistent with observations by \citet{manzano2024uncertainty}, who report that identifiability is strongest at small noise scales and degrades as noise increases.

\begin{figure}
    \centering
    \includegraphics[width=\linewidth]{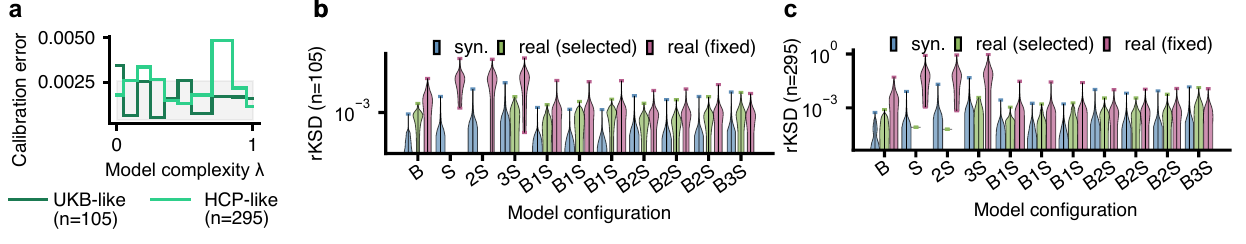}
    \caption{\textbf{Calibration for dMRI data.}
    \textbf{(a)} Model posterior calibration in terms of SBC across $\lambda$ for UKB- and HCP-like data.
\textbf{(b)} rKSD on the UKB dataset for synthetic data (true model) and real data, comparing a \emph{fixed} model to the model \emph{selected} by the model-posterior network (same as Fig.~\ref{fig:eval_b3s}c).
\textbf{(c)} Same as (b) for the HCP dataset.
    }
    \label{fig:app:dMRI_calibration_rKSD}
\end{figure}

\begin{table}
\caption{\textbf{Parameter inference performance metrics.} Median and 10th/90th quantiles of performance metrics on synthetic and real data for the effective sample size (ESS), the Kernel Stein discrepancy (KSD) and the predictive rooted mean square error (RMSE). Obtained via the tailored B3S model. Synthetic is over the whole model space, where real is restricted to specific models.}
\centering
\setlength{\tabcolsep}{3pt}
\renewcommand{\arraystretch}{1.05}
\resizebox{0.92\textwidth}{!}{%
\begin{tabular}{lccccc}
\hline
 & \multicolumn{2}{c}{Synthetic} & \multicolumn{3}{c}{Real (UKB)} \\
Metric & UKB-like & HCP-like & B1S & B2S & B3S \\
\hline
ESS  & $0.615\,[0.098,\,0.962]$ & $0.578\,[0.084,\,0.979]$ & $0.177\,[0.021,\,0.695]$ & $0.384\,[0.053,\,0.685]$ & $0.276\,[0.070,\,0.621]$ \\
KSD  & $0.304\,[0.000,\,11.753]$ & $0.141\,[0.000,\,11.867]$ & $0.306\,[0.000,\,1.546]$ & $0.389\,[0.000,\,1.390]$ & $0.062\,[0.000,\,0.758]$ \\
RMSE & $0.020\,[0.012,\,0.076]$ & $0.020\,[0.012,\,0.076]$ & $0.030\,[0.019,\,0.054]$ & $0.023\,[0.015,\,0.034]$ & $0.020\,[0.014,\,0.032]$ \\
\hline
\end{tabular}%
}
\label{tab:appendix_b3s_performance}
\end{table}

\begin{figure}
    \centering
    \includegraphics[width=\linewidth]{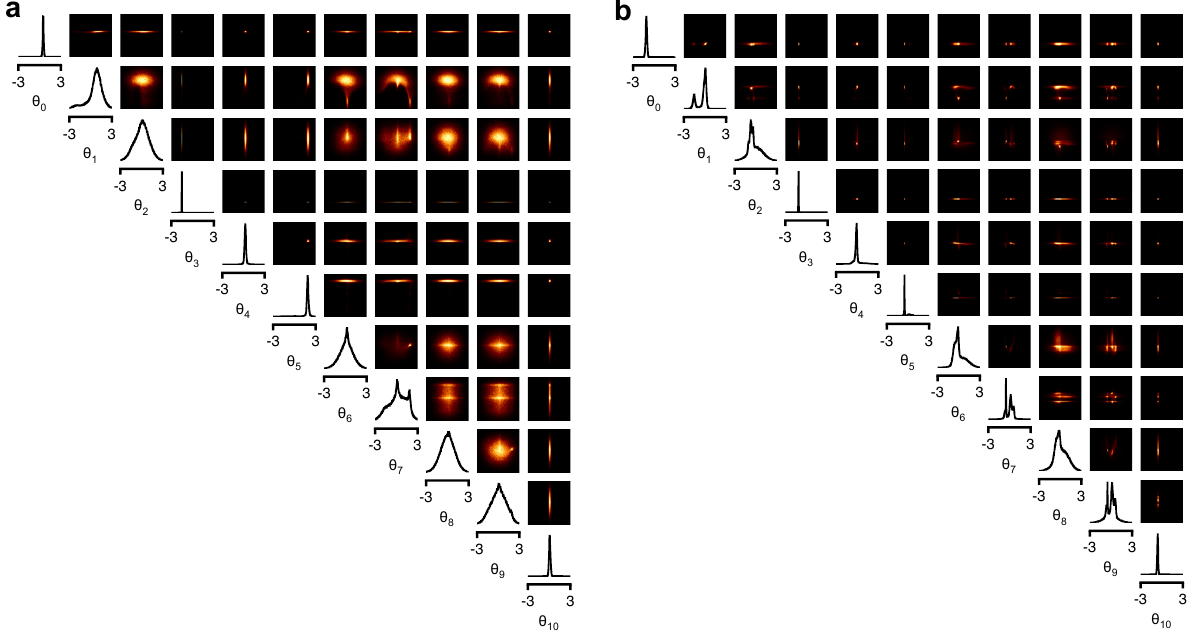}
    \caption{\textbf{Parameter posterior for Ball-and-Stick model.} Full parameter posterior distribution conditioned on the B3S model
    \textbf{(a)} for V1
    \textbf{(b)} for V2 as shown  in Fig.~\ref{fig:eval_b3s}.
    }
    \label{fig:b3s_pairplot}
\end{figure}

\begin{figure}
    \centering
    \includegraphics[width=\linewidth]{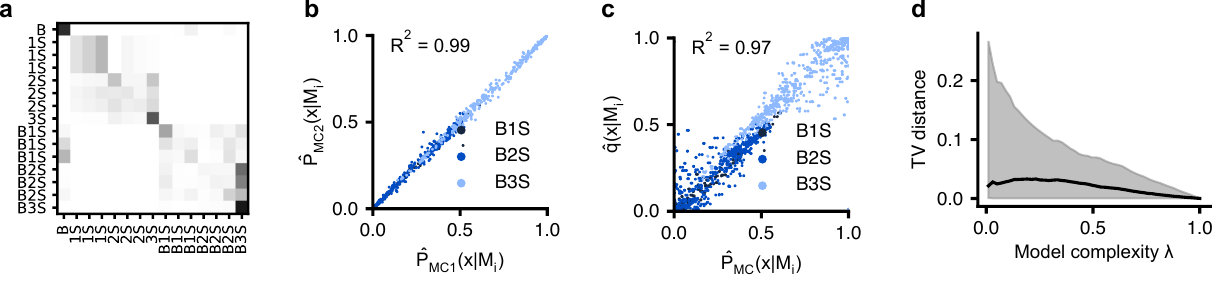}
    \caption{\textbf{B3S model selection evaluation.
    (a)} Confusion matrix for all models. \textbf{(b)} Alignment between two Monte Carlo (MC) approximations of the evidence.
    \textbf{(c)} Alignment between the approximated evidence obtained from the \ours\ model posterior probabilities compared to MC approximations.
    \textbf{(d)} Total variation (TV) distance across different model complexities $\lambda$.
    }
    \label{fig:b3s_model_selection_eval}
\end{figure}


\subsubsection{Extended model space inference}

While Ball 3 Stick model space is relatively small, we can extend it by adding alternative compartments (see Sec.~\ref{app:dMRI}, Tab.~\ref{tab:dmri_models_priors_params} for a table of considered compartments). Specifically a 'Stick' (S) can be considered a degenerate (rank 1) special case of a 'Tensor' (T) ~\citep{basser1994dti} (or of a 'Zeppelin' (Z) which itself is a constrained 'Tensor'). We consider the Ball-Stick-Zeppelin-Tensor (BSZT) space which contain three times a SZT components (ie. analogous to B3S). We further extend this spaces to BSZT-conv which includes additional componenents that are based on spherical convolution~\citep{canales2015spherical}. See \citet{fick2019dmipy} for an overview.

We investigate the performance of models trained on the BSZT and BSZT+conv spaces (see Sec.~\ref{sec:bszt_task_configuration} for details). In contrast to B3S, which can be evaluated using relatively standard approaches (including classification-based analyses~\citep{manzano2024uncertainty}), these model spaces are substantially larger, which makes evaluation more challenging. Additionally, while smaller models can be accurately estimated with the investigated data (i.e. are relatively well-constrained up to symmetries), larger models such as B3T are under constrained. This can also be seen at the associated posteriors which show high degree of parameter degeneracy (Fig.~\ref{fig:all_pairplot}).

We first consider simulation-based calibration. Overall, both the model and parameter predictions remain well-calibrated (Fig.~\ref{fig:appendix_all_calibration_error}a,b, left), although the calibration error is now slightly higher than the level expected under perfect calibration. For the BSZT space, deviations from the expected behavior are largely confined to specific settings (notably HCP acquisitions and a subset of models), whereas the BSZT+conv space exhibits more systematic deviations. We attribute this loss in performance also to computational constraints: simulating dMRI models based on spherical convolution is substantially more expensive (even with an additional GPU), which limits the effective coverage of the enlarged model space during training. We further corroborate these trends using RMSE, ESS, and KSD (Tab.~\ref{tab:appendix_all_performance}, ~\ref{tab:appendix_all_conv_performance}).

We additionally evaluate the model posterior network from a classification perspective on a randomly sampled model subspace (Fig.~\ref{fig:confusion_matrix_all_all_conv}). This is inherently a difficult task for two reasons. First, many ``complex'' models can represent ``simpler'' ones by setting the corresponding fraction parameters to (near) zero. Second, in this configuration we include genuinely equivalent models (since we removed informative fraction priors), and, more broadly, Stick, Zeppelin, and Ball components can be viewed as constrained parameterizations of the Tensor model.

Despite these challenges, the confusion matrices remain largely concentrated along the diagonal, with off-diagonal mass primarily distributed among equivalent or closely related models, as expected. The top-5 accuracy is \(94\%\) for the BSZT space and \(98\%\) for the BSZT+conv space (Tab.~\ref{tab:appendic_dmri_classifications_metrics_merged}; although the latter increase can be primarily attributed to increase class diversity within the subspace). We emphasize that this improvement should not be interpreted as a strict dominance of BSZT+conv; rather, it is at least partly explained by differences in the evaluated subset, which is more diverse in the BSZT+conv case and therefore can be easier to separate in a top-\(k\) sense.

\begin{figure}
    \centering
    \includegraphics[width=\linewidth]{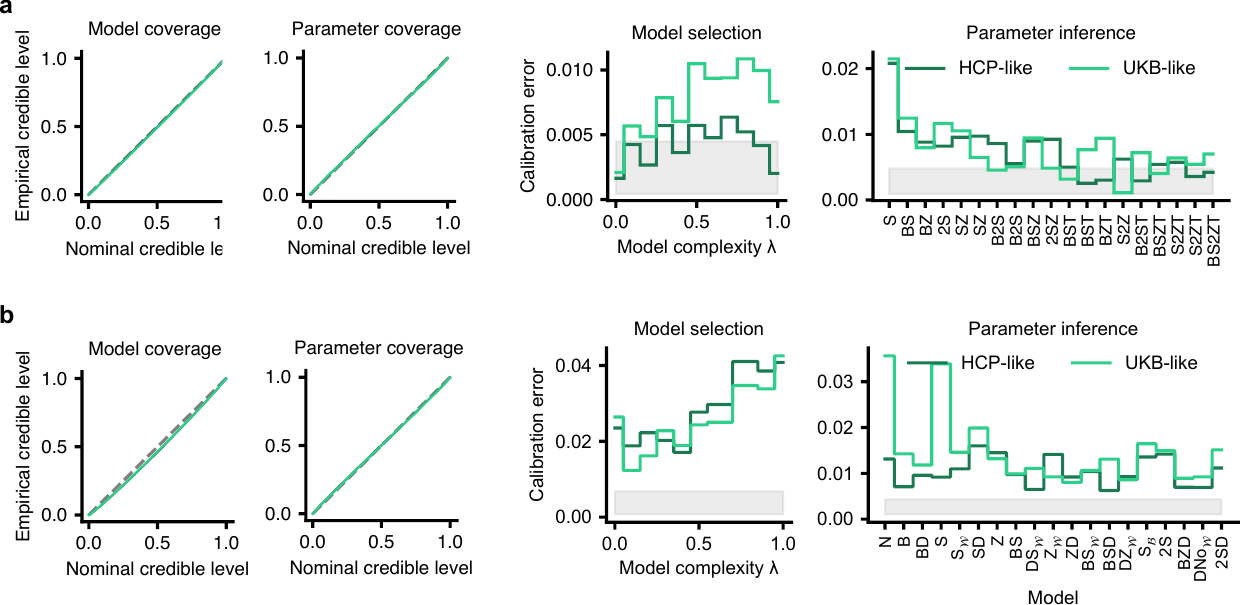}
    \caption{\textbf{SBC evaluation for extended dMRI model spaces.} Calibration plot and error (over $\lambda$ and models) for the model trained on
   \textbf{(a}) the BSZT  and
   \textbf{(b)} BSZT + conv space.}
    \label{fig:appendix_all_calibration_error}
\end{figure}

\begin{table}[ht]
\caption{\textbf{Parameter inference performance metrics.} Median and 10th/90th quantiles of performance metrics on synthetic and real data for the effective sample size (ESS), the Kernel Stein discrepancy (KSD) and the predictive rooted mean square error (RMSE) for the BSZT space model. Synthetic is over the whole model space, where real is restricted to specific models.}
\centering
\setlength{\tabcolsep}{3pt}
\renewcommand{\arraystretch}{1.05}
\resizebox{0.92\textwidth}{!}{%
\begin{tabular}{lccccc}
\hline
 & \multicolumn{2}{c}{Synthetic} & \multicolumn{3}{c}{Real (UKB)} \\
Metric & UKB-like & HCP-like & B3S & B3Z & B3T \\
\hline
ESS  & $0.109\,[0.026,\,0.962]$ & $0.081\,[0.028,\,0.986]$& $0.133\,[0.045,\,0.302]$ & $0.131\,[0.040,\,0.398]$ & $0.165\,[0.061,\,0.366]$ \\
KSD  & $1.144\,[0.000,\,7.086]$ & $1.776\,[0.000,\,10.411]$ & $0.004\,[0.000,\,0.107]$ & $0.000\,[0.000,\,0.014]$ & $0.000\,[0.000,\,0.000]$ \\
RMSE & $0.019\,[0.011,\,0.072]$ & $0.019\,[0.011,\,0.072]$ & $0.020\,[0.012,\,0.032]$ & $0.020\,[0.012,\,0.032]$ & $0.018\,[0.012,\,0.030]$ \\
\hline
\end{tabular}%
}
\label{tab:appendix_all_performance}
\end{table}

\begin{table}[ht]
\caption{\textbf{Parameter inference performance metrics.} Median and 10th/90th quantiles of performance metrics on synthetic and real data for the effective sample size (ESS), the Kernel Stein discrepancy (KSD) and the predictive rooted mean square error (RMSE) for the BSZT + conv model. Synthetic is over the whole model space, where real is restricted to specific models.}
\centering
\setlength{\tabcolsep}{3pt}
\renewcommand{\arraystretch}{1.05}
\resizebox{0.92\textwidth}{!}{%
\begin{tabular}{lccccc}
\hline
 & \multicolumn{2}{c}{Synthetic} & \multicolumn{3}{c}{Real (UKB)} \\
Metric & UKB-like & HCP-like & BS$_\mathcal{W}$S$_\mathcal{B}$ & No$_\mathcal{B}$ & 2TS$_\mathcal{W}$S$_\mathcal{B}$No$_\mathcal{W}$ \\
\hline
ESS  & $0.073\,[0.025,\,0.453]$  & $0.083\,[0.021,\,0.433]$  &  $0.109\,[0.028,\,0.464]$ & $0.053\,[0.020,\,0.335]$   & $0.135\,[0.043,\,0.410]$  \\
KSD  & $3.873\,[0.671,22.23]$  & $6.597\,[0.922,35.49]$  &  $0.001\,[0.000,\,0.016]$ & $1.000\,[0.089,\,4.306]$ & $0.000\,[0.000,\,0.000]$ \\
RMSE & $0.021\,[0.013,\,0.072]$ & $0.019\,[0.012\,0.073]$  & $0.022\,[0.013,\,0.049]$ & $0.028\,[0.014,\,0.065]$ & $0.018\,[0.012,\,0.031]$ \\
\hline
\end{tabular}%
}
\label{tab:appendix_all_conv_performance}
\end{table}

\begin{figure}
    \centering
    \includegraphics[width=\linewidth]{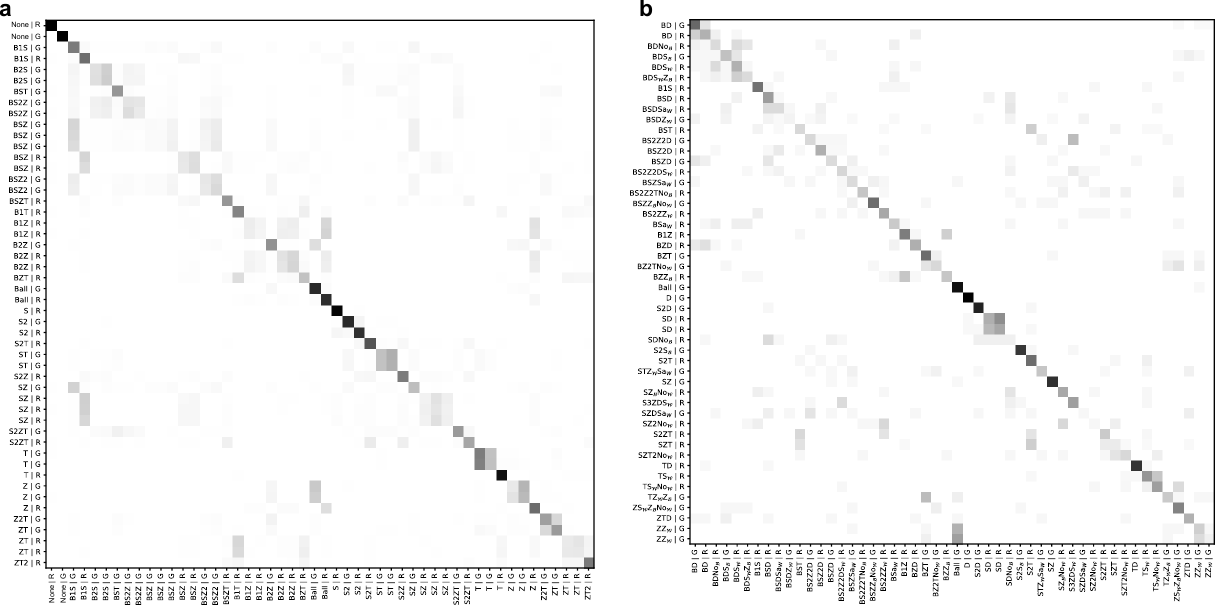}
    \caption{\textbf{Confusion matrices for \ours~ on dMRI data with full model family. }
    \textbf{(a)} on a 50 model subspace for the BSZT space model
    \textbf{(b)} on a 50 model subspace for the BSZT+conv space model}

    \label{fig:confusion_matrix_all_all_conv}
\end{figure}

\begin{figure}
    \centering
    \includegraphics[width=\linewidth]{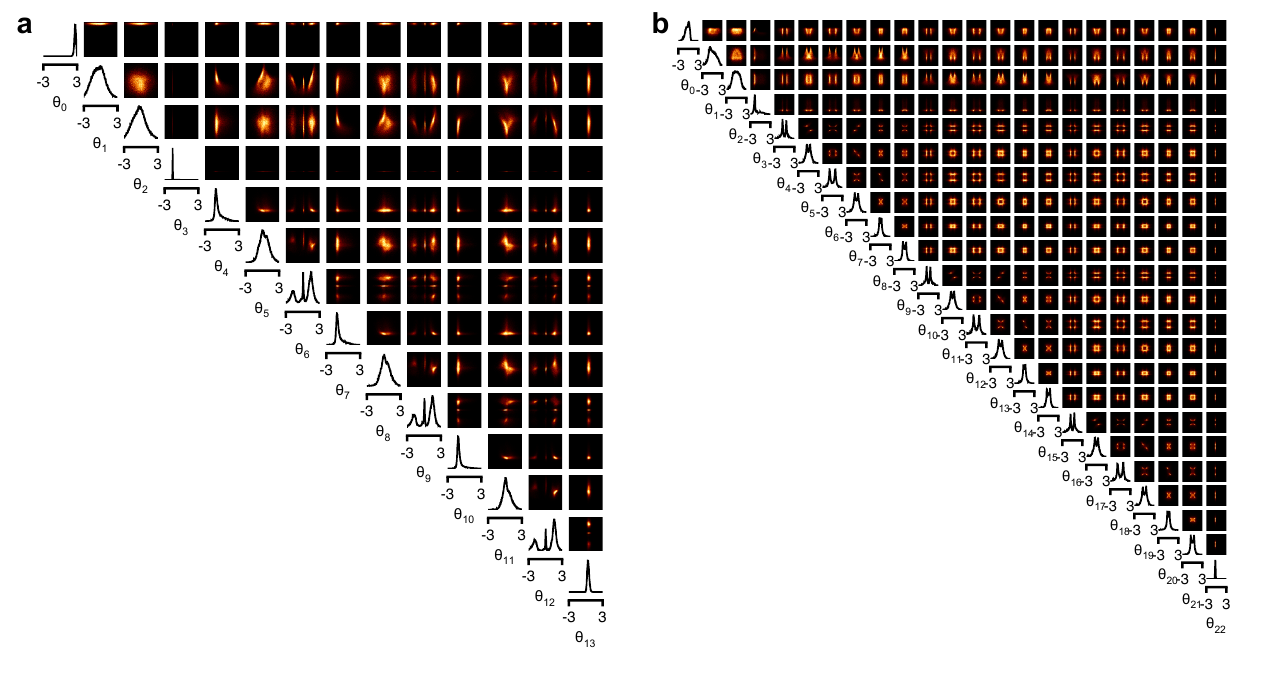}
    \caption{\textbf{Parameter posterior for dMRI.} Parameter posterior for the two voxels shown in Fig.~\ref{fig:eval_b3s} as estimated by the BSZT-space model.
    \textbf{(a)} Posterior for the B3S model (different from previous due to unshared diffusivity and uninformative fraction prior).
    \textbf{(b)} Posterior for the B3T model showing large degeneracy due to symmetries and nonidentifiability on current datasets. }
    \label{fig:all_pairplot}
\end{figure}

\subsubsection{dMRI FOD results}

Joint model--parameter inference assigns, for each voxel, both a model \(\vm\) and a corresponding parameter vector \(\vtheta_\vm\). This flexibility is a major strength, but it also complicates standard per-parameter analyses: parameters are model-specific, and their interpretation can change across models. For example, in B3S one can extract inferred diffusion ``orientations'' by inspecting the directions of the Stick compartments together with their associated fractions, but an analogous procedure is not directly comparable across other model classes.

To enable a consistent analysis across models, we instead map each posterior sample \((\vtheta_\vm, \vm)\) to a universal representation: the Orientation Distribution Function (ODF). The ODF is a spherical distribution that provides a common summary of the orientation content implied by any model in our family. We detail this conversion in Sec.~\ref{app:dMRI}. This representation allows us to visualize and compare joint (or model-averaged) posterior distributions in a single object that captures both the predicted orientation structure and its associated uncertainty.

For a slice of HCP data, we compare the fODF predicted by the baseline method Rumba~\citep{canales2015spherical} (maximum-likelihood spherical deconvolution with anatomical priors) to posterior-predictive fODFs obtained from B3S, B3T, and the BSZT-space model average. Overall, we observe substantial qualitative agreement in the dominant diffusion directions across methods on this dataset, while the associated uncertainty differs markedly between approaches. In particular, \ours~ yields uncertainty estimates that vary spatially and across configurations, reflecting both measurement noise and model ambiguity in a principled way.

A key limitation is that determining the ``most faithful'' fiber orientation in vivo is itself an open problem, and a definitive evaluation of orientation accuracy is beyond the scope of this work. Consequently, agreement or disagreement in fODF uncertainty should not be over-interpreted as direct evidence of greater anatomical correctness. What we can assess reliably, however, is statistical accuracy relative to the observed dMRI signal. Using this criterion, we find that posterior samples produced by \ours~ are better aligned with the observed data than those from the baseline methods, indicating that the inferred models are better supported by the measurements (Fig.~\ref{fig:eval_all}).

\begin{figure}
    \centering
    \includegraphics[width=0.9\linewidth]{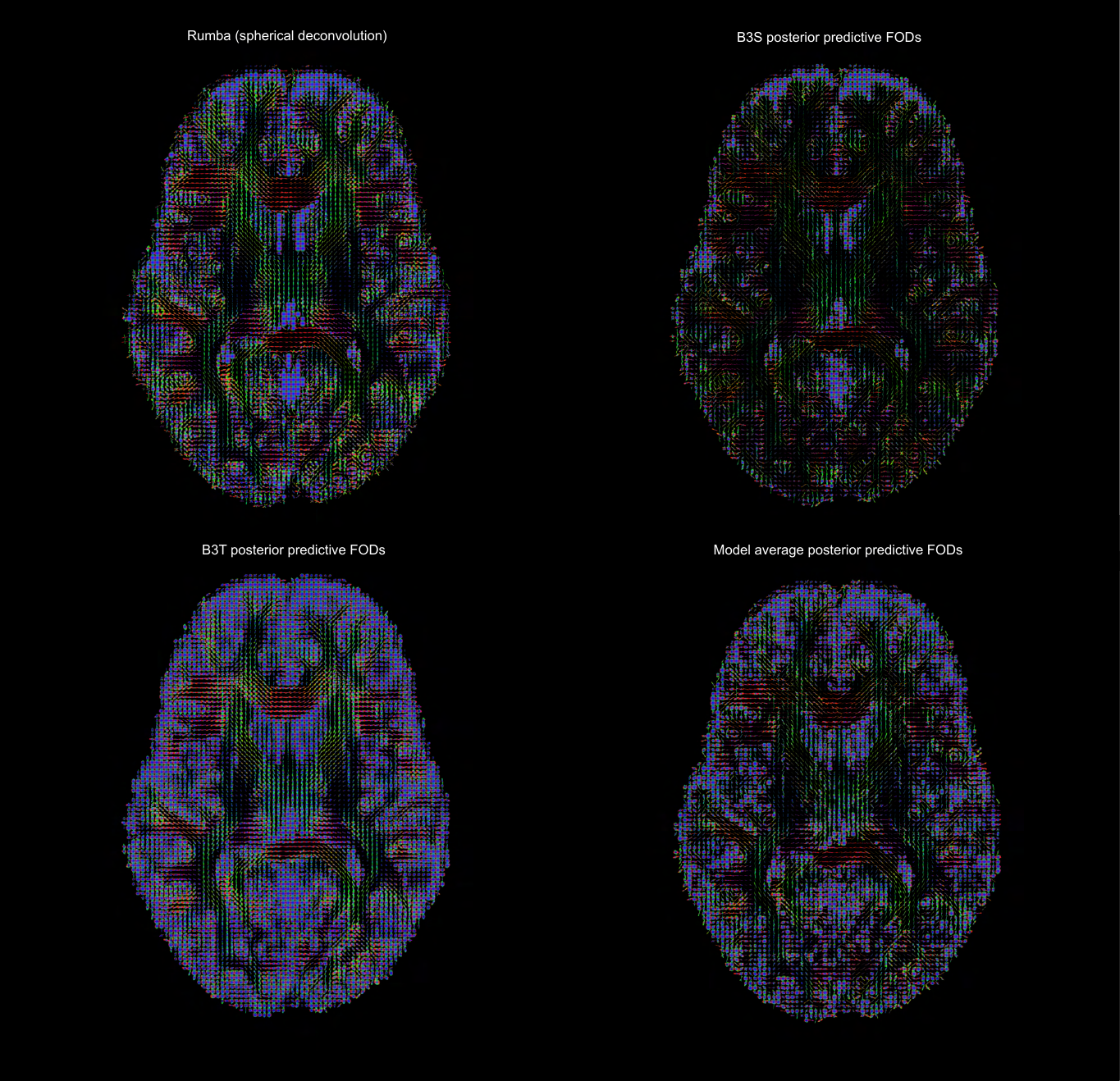}
    \caption{Posterior predictive fiber orientation distributions for a classical spherical deconvolution based method (Rumba, \citep{canales2015spherical}, as implemented in dipy \citep{garyfallidis2014dipy}), presuming the B3S model, the B3T or by considering the \textbf{global} posterior predictive FOD averaged over all possible tensor-based models.}
    \label{fig:appendix_fod}
\end{figure}


\begin{table}[h]
\centering
\caption{Classification metrics across BSZT spaces (macro-averaged).}
\begin{tabular}{lccc}
\hline
\textbf{Metric} & \textbf{B3S space} & \textbf{BSZT space} & \textbf{BSZT + conv space} \\
\hline
Accuracy (Top-1, Macro) & 0.351 & 0.428 & 0.422 \\
Accuracy (Top-5, Macro) & 0.933 & 0.940 & 0.980 \\
Macro precision          & 0.362 & 0.404 & 0.420 \\
Macro recall             & 0.351 & 0.428 & 0.422 \\
Macro F1                 & 0.337 & 0.403 & 0.408 \\
\hline
\end{tabular}
\label{tab:appendic_dmri_classifications_metrics_merged}
\end{table}

\subsubsection{Additional model selection results}
\label{sec:app_model_selection}




\label{sec:appendix_model_selection}
Model selection is performed on a (prior-weighted) evidential basis, i.e.\ by choosing the model with the largest marginal likelihood (Bayesian evidence). In practice, we consider two complementary procedures:

\begin{itemize}
    \item[(i)] \textit{Model selection (within a user-specified subspace).}
    Given a user-selected subspace \(\vM_{\text{sub}} \subset \vM\), we select
    \begin{equation}
        \vm^* \;=\; \arg\max_{\vm \in \vM_{\text{sub}}} q(\vm \mid \vx_o).
    \end{equation}
    For a uniform prior (\(\lambda = 1.0\)), this is equivalent to selecting the model with maximal Bayesian evidence. We solve this problem exactly by evaluating \(q(\vm \mid \vx_o)\) for every \(\vm \in \vM_{\text{sub}}\) and returning the maximizer.

    \item[(ii)] \textit{Model discovery followed by model selection.}
    We first solve the global optimization problem
    \begin{equation}
        \vm^* \;=\; \arg\max_{\vm \in \vM} q(\vm \mid \vx_o).
    \end{equation}
    Since this is a discrete optimization over a large model space, we approximate it by drawing 100 samples from the model posterior and selecting the model with the highest posterior probability among the sampled set. This ``discovery'' step can yield many distinct models across the brain; to obtain a compact and interpretable set, we restrict attention to the 10 most frequently selected models, denoted \(\vM_{\text{top10}}\). We then perform voxel-wise selection within \(\vM_{\text{top10}}\) using procedure (i).
\end{itemize}

As in the symbolic regression task, the model-complexity prior strongly influences selection. When restricting model comparison to the B(1/2/3)(S/Z/T) subset, more complex models become less probable as $\lambda$ decreases. However, the rate of this decline depends on the acquisition: for HCP data, complex models remain competitive over a wider range of $\lambda$ (Fig.~\ref{fig:app:model_selection}\textbf{a}).

Beyond such constrained comparisons, the learned model posterior enables exploratory analysis over the full combinatorial space and highlights high-probability component combinations. Notably, when selecting over the full BSZT space, combinations such as BZT, BT, or BZ consistently rank highest across all $\lambda$. This suggests that models that dominate within the restricted B(1/2/3)(S/Z/T) subset (e.g., B3T) can often be replaced by simpler multi-component alternatives once the search space is expanded (Fig.~\ref{fig:app:model_selection}\textbf{b}). Expanding further to the BSZT + conv space shifts the posterior mass again: Watson- or Bingham-stick variants combined with Balls or Tensor components (and NODDI-style formulations) become substantially more dominant (Fig.~\ref{fig:app:model_selection}\textbf{b}). These shifts underscore that model-selection outcomes are conditional on the candidate set: there may always exist a better model outside the chosen search space, so claims about the “best” model should be framed relative to the models considered.

Overall, the posterior concentrates on multi-compartment combinations that better explain the diffusion MRI signal, consistent with prior cross-validation studies \citep{panagiotaki2012compartment, ferizi2014ranking, ferizi2015white, ferizi2017diffusion}. However, detailed analysis remains for future work. While the selected models are consistent with observed data, to answer domain-specific question it's better to constrain the spaces/priors to compare "appropriate" models for the question at hand.

\begin{figure}[t]
    \centering
    \includegraphics[width=\linewidth]{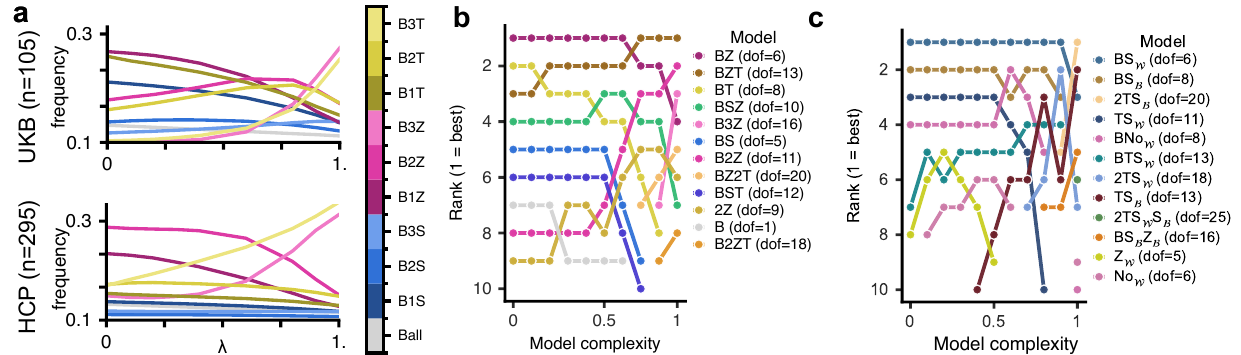}
    \caption{\textbf{Model selection results on extended model spaces} \textbf{a} Model selection frequencies if constrained to a set of B(1/2/3)(S/Z/T) models for the UKB and HCP dataset. \textbf{b)} Top 10 ranking of the whole BSZT space across lambda for the UKB dataset. \textbf{c)} Top 10 ranking of the BSZT+conv space across lambda for the UKB dataset. }
    \label{fig:app:model_selection}
\end{figure}


\clearpage
\section{Problem specifications}
\subsection{Symbolic regression problem specifications}
\label{sec:app_symbolic_regression_task}

We list all components in Tab. \ref{tab:extended_symbolic_components} as well as their naming. For $K=15$ we used exactly the setting as reported in \citet{schroder2024simultaneous} with noise model 'NoiseObserver','NoiseIncreasing', 'NoiseDecreasing', 'NoiseQuadratic', 'NoiseQuadraticDecreasing';otherwise, we used all noise models.

Note that we allow for repeated components. For all settings investigated, the symbolic base functions are given by Tab.~\ref{tab:basis_libraries}.
\begin{longtable}{p{0.18\linewidth} p{0.78\linewidth}}
\label{tab:basis_libraries}\\

\caption[]{Explicit listing of base function (see Tab.~\ref{tab:extended_symbolic_components} for definitions) in order i.e. mask index $i$ corresponds to the i'th base function listed.}\\
\toprule
\textbf{Setting} & \textbf{Basis functions} \\
\midrule
\endfirsthead

\toprule
\textbf{Setting} & \textbf{Basis functions} \\
\midrule
\endhead

\midrule
\multicolumn{2}{r}{\emph{Continued on next page}}\\
\endfoot

\bottomrule
\endlastfoot

\textbf{$K{=}15$}
& Linear, Linear, Quadratic, ShiftedSquare, Cubic, Sinusoidal, Cosinusoidal, ConstantWide, ConstantPositive, TanhRight, TanhLeft, GaussianBump, GaussianWide, RampUp, RampDown \\
\addlinespace

\textbf{$K{=}30$}
& Linear, Quadratic, ShiftedSquare, Cubic, Sinusoidal, Cosinusoidal, ConstantWide, ConstantPositive, TanhRight, TanhLeft, TanhCentered, GaussianBump, GaussianWide, RampUp, RampDown, QuarticScaled, QuinticScaled, SinusoidalPhase, CosinusoidalPhase, ExponentialDecay, SaturatingExponential, Logarithmic, SquareRoot, Reciprocal, AbsoluteValue, InverseQuadratic, Lorentzian, Sigmoid, DampedSinusoidal, DampedCosinusoidal \\
\addlinespace

\textbf{$K{=}50$}
& Linear, Quadratic, ShiftedSquare, Cubic, Sinusoidal, Cosinusoidal, ConstantWide, ConstantPositive, TanhRight, TanhLeft, TanhCentered, GaussianBump, GaussianWide, RampUp, RampDown, QuarticScaled, QuinticScaled, SinusoidalPhase, CosinusoidalPhase, ExponentialDecay, SaturatingExponential, Logarithmic, SquareRoot, Reciprocal, AbsoluteValue, InverseQuadratic, Lorentzian, Sigmoid, DampedSinusoidal, DampedCosinusoidal, ExponentialGrowth, PowerLawDecay, ArctangentStep, HyperbolicSecant, SincDecay, AbsoluteSinusoidal, RectifiedLinear, Softplus, Gompertz, LinearFractional, SineSquared, TriangularBump, Linear, Quadratic, ShiftedSquare, Cubic, Sinusoidal, Cosinusoidal, ConstantWide, ConstantPositive \\
\addlinespace

\textbf{$K{=}80$}
& Linear, Quadratic, ShiftedSquare, Cubic, Sinusoidal, Cosinusoidal, ConstantWide, ConstantPositive, TanhRight, TanhLeft, TanhCentered, GaussianBump, GaussianWide, RampUp, RampDown, QuarticScaled, QuinticScaled, SinusoidalPhase, CosinusoidalPhase, ExponentialDecay, SaturatingExponential, Logarithmic, SquareRoot, Reciprocal, AbsoluteValue, InverseQuadratic, Lorentzian, Sigmoid, DampedSinusoidal, DampedCosinusoidal, ExponentialGrowth, PowerLawDecay, ArctangentStep, HyperbolicSecant, SincDecay, AbsoluteSinusoidal, RectifiedLinear, Softplus, Gompertz, LinearFractional, SineSquared, TriangularBump, Linear, Quadratic, ShiftedSquare, Cubic, Sinusoidal, Cosinusoidal, ConstantWide, ConstantPositive, TanhRight, TanhLeft, TanhCentered, GaussianBump, GaussianWide, RampUp, RampDown, QuarticScaled, QuinticScaled, SinusoidalPhase, CosinusoidalPhase, ExponentialDecay, SaturatingExponential, Logarithmic, SquareRoot, Reciprocal, AbsoluteValue, InverseQuadratic, Lorentzian, Sigmoid, DampedSinusoidal, DampedCosinusoidal, ExponentialGrowth, PowerLawDecay, ArctangentStep, HyperbolicSecant, SincDecay, AbsoluteSinusoidal, RectifiedLinear, Softplus \\
\addlinespace

\textbf{$K{=}100$}
& Linear, Quadratic, ShiftedSquare, Cubic, Sinusoidal, Cosinusoidal, ConstantWide, ConstantPositive, TanhRight, TanhLeft, TanhCentered, GaussianBump, GaussianWide, RampUp, RampDown, QuarticScaled, QuinticScaled, SinusoidalPhase, CosinusoidalPhase, ExponentialDecay, SaturatingExponential, Logarithmic, SquareRoot, Reciprocal, AbsoluteValue, InverseQuadratic, Lorentzian, Sigmoid, DampedSinusoidal, DampedCosinusoidal, ExponentialGrowth, PowerLawDecay, ArctangentStep, HyperbolicSecant, SincDecay, AbsoluteSinusoidal, RectifiedLinear, Softplus, Gompertz, LinearFractional, SineSquared, TriangularBump, Linear, Quadratic, ShiftedSquare, Cubic, Sinusoidal, Cosinusoidal, ConstantWide, ConstantPositive, TanhRight, TanhLeft, TanhCentered, GaussianBump, GaussianWide, RampUp, RampDown, QuarticScaled, QuinticScaled, SinusoidalPhase, CosinusoidalPhase, ExponentialDecay, SaturatingExponential, Logarithmic, SquareRoot, Reciprocal, AbsoluteValue, InverseQuadratic, Lorentzian, Sigmoid, DampedSinusoidal, DampedCosinusoidal, ExponentialGrowth, PowerLawDecay, ArctangentStep, HyperbolicSecant, SincDecay, AbsoluteSinusoidal, RectifiedLinear, Softplus, Gompertz, LinearFractional, SineSquared, TriangularBump, Linear, Quadratic, ShiftedSquare, Cubic, Sinusoidal, Cosinusoidal, ConstantWide, ConstantPositive, TanhRight, TanhLeft, TanhCentered, GaussianBump, GaussianWide, RampUp, RampDown, QuarticScaled \\
\end{longtable}

\newcommand{\Uniform}{\mathcal{U}}
\newcommand{\Dirichlet}{\mathrm{Dirichlet}}
\newcommand{\HalfNormal}{\mathrm{HalfNormal}}

\subsection{dMRI problem specifications}
\label{app:dMRI}

\subsubsection{Background}

Diffusion MRI (dMRI) modelling is an \emph{inverse problem}: given diffusion-encoding settings and noisy magnitude measurements, we seek latent tissue parameters that determine the expected signal attenuation. In the pulsed-gradient spin-echo (PGSE) experiment, the encoding is conveniently summarized by the $b$-tensor (or, in the common single-direction case, by a scalar $b$ and a unit direction $\mathbf{g}\in\mathbb{S}^2$) \cite{stejskal1965spin,mattiello1997bmatrix}.

A useful starting point is to model water motion as Brownian diffusion. If spin displacements over the effective diffusion time satisfy a Gaussian law (as for free Brownian motion, or as an approximation for \emph{hindered} diffusion in tissue at the voxel scale), then the displacement $\Delta\mathbf{x}$ has density
\[
p(\Delta\mathbf{x})=\mathcal{N}\!\big(\mathbf{0},\,2\Delta\,\mathbf{D}\big),
\]
with diffusion time $\Delta$ and a positive definite diffusion tensor $\mathbf{D}\succ 0$. In PGSE, the normalized signal can be written (under narrow-pulse and Gaussian phase approximations) as the characteristic function of the displacement distribution evaluated at the encoding wavevector,
\[
\frac{S}{S_0}\approx \mathbb{E}\!\left[e^{i\,\mathbf{q}^\top \Delta\mathbf{x}}\right].
\]
For Gaussian $\Delta\mathbf{x}$ this expectation is available in closed form and yields a mono-exponential attenuation in the encoding strength. In the single-direction case one obtains
\[
\frac{S(b,\mathbf{g})}{S_0}\approx \exp\!\big(-b\,D_{\mathrm{app}}(\mathbf{g})\big),
\qquad
D_{\mathrm{app}}(\mathbf{g})=\mathbf{g}^\top \mathbf{D}\,\mathbf{g}.
\]
This second-order parameterization defines diffusion tensor imaging (DTI) \cite{basser1994dti},
\[
\frac{S(b,\mathbf{g})}{S_0}=\exp\!\left(-b\,\mathbf{g}^{\top}\mathbf{D}\mathbf{g}\right),
\qquad \mathbf{D}\succ 0.
\]
Mathematically, DTI assumes that within each voxel the ensemble displacement is adequately summarized by a \emph{single} multivariate Gaussian (equivalently, by its covariance), so that direction-dependent signal decay is fully captured by the quadratic form $\mathbf{g}^\top\mathbf{D}\mathbf{g}$. DTI provides a low-dimensional description of anisotropy and a dominant local orientation, but it is not expressive enough when multiple micro-environments and/or multiple fiber orientations contribute within the same voxel, in which case the voxel-scale displacement distribution is better described as non-Gaussian or multi-compartment \cite{alexander2019microstructure,novikov2021present}.

\paragraph{Orientation distributions and spherical convolution.}
Many forward models describe a \emph{single-fiber} response conditional on an orientation $\mathbf{n}\in\mathbb{S}^2$. In the same spirit as DTI---where hindered diffusion within a voxel is approximated by a single Gaussian displacement (equivalently, a single diffusion tensor)---a single-fiber kernel can be seen as a locally oriented response. If a voxel contains dispersion and/or multiple fiber populations, a single Gaussian (single tensor) is no longer adequate, and the signal is better modeled as an average over orientations. This is achieved by integrating the single-fiber kernel against an orientation distribution function (ODF). When the kernel depends on $\mathbf{n}$ only through the inner product $\mathbf{g}^{\top}\mathbf{n}$, this average becomes a spherical convolution:
\[
g(b,\mathbf{g})
=
\int_{\mathbb{S}^2}
g_{\mathrm{sf}}\!\big(b,\mathbf{g}^{\top}\mathbf{n}\mid \theta\big)\,
p(\mathbf{n}\mid \phi)\,
d\mathbf{n},
\]
with a parametric ODF $p(\mathbf{n}\mid \phi)$ (e.g.\ Watson or Bingham) or a nonparametric ODF as in spherical deconvolution \cite{tournier2007csd}. Dispersion-aware models can improve orientation estimates, but add degrees of freedom and can increase sensitivity to acquisition design and regularization \cite{behrens2007multifibre,jbabdi2012model}.

\paragraph{Why model choice matters.}
In practice, model choice should be guided by anatomy and acquisition. The compartment family and priors should reflect plausible voxel-scale structure (e.g.\ single bundle vs.\ crossings; dispersed vs.\ tightly aligned orientations; presence/absence of free-water-like signal) and be supported by the available $b$-values, SNR, and gradient sampling. Without such constraints, flexible mixtures may fit the signal while implying implausible tissue configurations, which can propagate to downstream fiber reconstruction and tractography \cite{thomas2014anatomical,ferizi2014ranking}. Accordingly, it is often useful to constrain inference via explicit priors, structured model selection, and sensitivity checks across plausible compartment families \cite{panagiotaki2012compartment,alexander2019microstructure,jelescu2017designvalidation}. Toolkits such as \texttt{DMIPY} expose these kernels as composable modules, enabling rapid specification of multi-compartment mixtures, dispersed kernels, and acquisition-dependent forward operators \cite{fick2019dmipy}. Other libraries such as \texttt{DIPY} \citep{garyfallidis2014dipy} provide a collection of established, easy-to-use pipelines (with limited but practical tuning, e.g.\ tissue-specific options for white vs.\ grey matter). Widely used packages such as \texttt{FSL} provide end-to-end workflows for diffusion processing and modelling. In particular, \texttt{BedpostX} performs Bayesian estimation via MCMC and uses shrinkage priors (ARD) to discourage unsupported fibre populations, helping stabilize inference under typical acquisitions.

\subsubsection{Implementation details (Table~\ref{tab:dmri_models_priors_params}).}
\label{sec:dmri_model_implemenation_details}
To connect the components in Table~\ref{tab:dmri_models_priors_params} within a single generative model, we use a \emph{multi-compartment} formulation in which the normalized diffusion signal is a convex mixture of $K{+}1$ candidate signal compartments and one noise model selected from a small family:
\begin{equation}
\frac{S(b,\vec{\mathbf{b}} \mid \vTheta)}{S_0}
= \sum_{k=0}^{K} M_k \, f_k \, g_k\!\left(b,\vec{\mathbf{b}}\mid\vtheta_k\right)
\; \qquad S_o \sim p\left(\frac{S(b,\vec{\mathbf{b}} \mid \vTheta)}{S_0} \mid \widetilde{M}_m, \vtheta_m \right)
\label{eq:multicompartment}
\end{equation}
Each $g_k(b,\vec{\mathbf{b}}\mid\vtheta_k)$ is a compartment kernel from the library in Table~\ref{tab:dmri_models_priors_params}
(e.g.\ Ball, Stick, Zeppelin, Tensor, and dispersed/convolutional variants such as Watson/Bingham sticks and zeppelins), with compartment-specific parameters $\vtheta_k$ (diffusivities, orientations, dispersion parameters, etc.). The weights $\mathbf{f}=(f_0,\ldots,f_K)$ are \emph{fractions} constrained to the simplex $f_k \ge 0, \sum_{k=0}^{K} f_k M_k = 1 $ which we encode using a Dirichlet prior, $\mathbf{f}\sim\Dirichlet(\boldsymbol{\alpha})$.

The binary indicators $M_k\in\{0,1\}$ implement \emph{structural model selection} by turning candidate compartments on/off, while the noise indicators $\widetilde{M}_m\in\{0,1\}$ select among candidate noise models $\epsilon_m$ (e.g.\ Gaussian or Rician magnitude noise; Table~\ref{tab:dmri_models_priors_params}). Noise models are treated as \emph{mutually exclusive}; we enforce $\sum_{m=0}^{M-1}\widetilde{M}_m=1$ so that exactly one noise model is active. Collecting all unknowns yields
$
\vTheta=\Big(\mathbf{f},\{M_k,\vtheta_k\}_{k=0}^{K},\{\widetilde{M}_m,\vtheta^{(\epsilon)}_m\}_{m=0}^{M-1}\Big),
$
and the priors over $\vtheta_k$ and $\vtheta^{(\epsilon)}_m$ are taken directly from Table~\ref{tab:dmri_models_priors_params}.

\paragraph{Parametric spherical convolution.}
We implement two parametric representations of orientation distribution functions (ODFs) on $\mathbb{S}^2$ \citep{watson1965equatorial,bingham1974antipodally}. The Watson distribution (axially symmetric, antipodally symmetric) is
\begin{equation}
f_{\mathcal{W}}(\mathbf{n}\mid \boldsymbol{\mu},\kappa)
=
C_{\mathcal{W}}(\kappa)\exp\!\big(\kappa(\boldsymbol{\mu}^\top\mathbf{n})^2\big),
\qquad
\mathbf{n}\in\mathbb{S}^2,
\qquad
C_{\mathcal{W}}(\kappa)=\frac{1}{4\pi\,{}_1F_1\!\left(\tfrac12;\tfrac32;\kappa\right)}.
\label{eq:watson_short}
\end{equation}
It is parameterized by $\boldsymbol{\mu}\in\mathbb{S}^2$ and $\kappa\in\mathbb{R}$ (3 degrees of freedom: 2 for $\boldsymbol{\mu}$ and 1 for $\kappa$). The Bingham distribution (elliptical, antipodally symmetric), as implemented in \citet{fick2019dmipy}, is
\begin{equation}
f_{\mathcal{B}}(\mathbf{n}\mid \mathbf{A})
=
C_{\mathcal{B}}(\mathbf{A})\exp\!\big(\mathbf{n}^\top\mathbf{A}\mathbf{n}\big),
\qquad
\mathbf{n}\in\mathbb{S}^2,
\qquad
\mathbf{A}=\kappa\,\boldsymbol{\mu}\boldsymbol{\mu}^\top+\beta\,\boldsymbol{\mu}_\beta\boldsymbol{\mu}_\beta^\top,
\label{eq:bingham_short}
\end{equation}
with axes $\boldsymbol{\mu},\boldsymbol{\mu}_\beta\in\mathbb{S}^2$ and concentrations $\kappa,\beta\in\mathbb{R}$. The normalizer $C_{\mathcal{B}}(\mathbf{A})$ has no closed form and is approximated via numerical integration.

To implement spherical convolutions with a fixed single-fiber response kernel---as required by dispersion/convolutional components (e.g.\ NODDI-style formulations)---we compute convolutions in the spherical-harmonics (SH) basis \citep{garyfallidis2014dipy,fick2019dmipy}. We approximate an ODF by a truncated SH expansion
$f(\mathbf{n})\approx \sum_{\ell=0}^{L}\sum_{m=-\ell}^{\ell} c_{\ell m}\,Y_{\ell m}(\mathbf{n})$
and represent an axially symmetric response kernel $R(\mathbf{n})$ by its zonal SH coefficients $\{r_\ell\}_\ell$ (equivalently, a Legendre series). In this representation, spherical convolution reduces to per-band multiplication,
$\tilde{c}_{\ell m}=r_\ell\,c_{\ell m}$,
which is more efficient and numerically stable than direct quadrature over $\mathbb{S}^2$ \citep{garyfallidis2014dipy,fick2019dmipy}. In practice, we (i) project the Watson/Bingham ODF onto SH coefficients (truncated to $L=13$), (ii) apply the coefficient-wise multiplication by $\{r_\ell\}$, and (iii) reconstruct the convolved angular profile on the required gradient directions.

\paragraph{Conversion to ODF.}
To compare fiber-orientation uncertainty across heterogeneous model configurations, we map each fitted model to an orientation distribution function (ODF), denoted by ${\mathrm{ODF}}(\mathbf{n})$, whenever possible. For models already defined via an ODF (or an ODF convolved with a response kernel), this mapping is immediate. For tensor-based compartments, we use the analytical single-tensor ODF of \citet{aganj2010reconstruction}. Given a diffusion tensor $\mathbf{D}=\mathbf{R}\,\mathrm{diag}(\boldsymbol{\lambda})\,\mathbf{R}^\top$ with eigenvalues $\boldsymbol{\lambda}=(\lambda_1,\lambda_2,\lambda_3)$ and eigenvectors $\mathbf{R}\in\mathbb{R}^{3\times 3}$ (columns), the induced ODF on directions $\mathbf{n}\in\mathbb{S}^2$ is
\begin{equation}
\mathrm{ODF}_{T}(\mathbf{n} \mid \mathbf{D})
=
\frac{1}{4\pi\,\sqrt{\det(\mathbf{D})}\,\big(\mathbf{n}^\top \mathbf{D}^{-1}\mathbf{n}\big)^{3/2}}
=
\frac{1}{4\pi\,\sqrt{\lambda_1\lambda_2\lambda_3}\,\big(\mathbf{n}^\top \mathbf{D}^{-1}\mathbf{n}\big)^{3/2}}.
\label{eq:single_tensor_odf}
\end{equation}
This includes the axially symmetric \emph{Zeppelin} tensor by setting $\boldsymbol{\lambda}=(\lambda_{\parallel},\lambda_{\perp},\lambda_{\perp})$ and choosing $\mathbf{R}$ such that its principal axis aligns with $\boldsymbol{\mu}$:
\begin{equation}
\mathrm{ODF}_{Z}(\mathbf{n} \mid \boldsymbol{\mu},\lambda_{\parallel},\lambda_{\perp})
=
\frac{1}{4\pi\,\sqrt{\lambda_{\parallel}\lambda_{\perp}^2}\,\big(\mathbf{n}^\top \mathbf{D}_{\mathcal{Z}}^{-1}\mathbf{n}\big)^{3/2}},
\qquad
\mathbf{D}_{\mathcal{Z}}=\mathbf{R}(\boldsymbol{\mu})\,\mathrm{diag}(\lambda_{\parallel},\lambda_{\perp},\lambda_{\perp})\,\mathbf{R}(\boldsymbol{\mu})^\top.
\label{eq:zeppelin_odf}
\end{equation}
The \emph{Stick} arises as the limiting case $\lambda_{\perp}\to 0$ (rank-1), yielding an ODF concentrated on a single axis (in the distributional sense),
$$\mathrm{ODF}_{S}(\mathbf{n} \mid \boldsymbol{\mu})\propto \delta(\mathbf{n};\boldsymbol{\mu})+\delta(\mathbf{n};-\boldsymbol{\mu}).$$

Any model configuration can thus be represented in ODF-space as a mixture, with mixture weights given by the fractions $\mathbf{f}$:
$\mathrm{ODF}(\mathbf{n} \mid \mathcal{M},\vTheta)=\sum_{k=0}^{K} f_k\,M_k\,\mathrm{ODF}_{\mathcal{M}_k}(\mathbf{n}\mid \vtheta_k)$.
Given a posterior over $(\mathcal{M},\vTheta)$, we define the posterior predictive ODF in local (model-conditional) and global (model-averaged) form as
$\mathrm{ODF}_{\mathrm{local}}(\mathbf{n}\mid \mathcal{M}_k,\mathbf{x}_o)=\mathbb{E}_{p(\vtheta_k\mid \mathcal{M}_k,\mathbf{x}_o)}[\mathrm{ODF}_{\mathcal{M}_k}(\mathbf{n}\mid \vtheta_k)]$
and
$\mathrm{ODF}_{\mathrm{global}}(\mathbf{n}\mid \mathbf{x}_o)=\mathbb{E}_{p(\mathcal{M},\vTheta\mid \mathbf{x}_o)}[\mathrm{ODF}(\mathbf{n}\mid \mathcal{M},\vTheta)]$.

\paragraph{Model Prior}
Because candidate components differ substantially in parameter count, we quantify model complexity by the number of parameters rather than by the number of components. We use a dimension-penalized prior over model indicators
$$p(\vm \mid \lambda)=\prod_k \mathrm{Ber}\big(M_k, p_k(\lambda,\text{dim}(\vtheta_{\mathcal{M}_k}) )\big),$$
with $\text{logit}(p_k)=\text{logit}(p_0)-(1-\lambda)\cdot \lambda_{\text{max}}\cdot \text{dim}(\vtheta_{\mathcal{M}_k})$ and $p_k=\sigma(\text{logit}(p_k))$.
Therefore $p_k$ decreases with increasing parameter dimension of $\vtheta_{\mathcal{M}_k}$.
We draw ${\lambda}\sim \mathrm{Unif}([0,1])$ and set $\lambda_{\text{max}}=4$ and $p_0=0.5$.

\subsubsection{Random acquisition protocol generation}
We generated random \texttt{bvals} and \texttt{bvecs} as follows:
\begin{itemize}
    \item[(i)] \textbf{UKB-like}: We fixed $n=105$ acquisitions. For each acquisition, we sampled the $b$-value from a mixture of (a) a fixed ``typical'' UKB-like set $\{b_i^{\mathrm{typ}}\}_{i=1}^{n}$ and (b) a uniform draw $b_i^{\mathrm{rand}}\sim \mathrm{Unif}(0,6000)$, with mixture weights $(0.5,0.5)$. We then applied a small dropout mask $m_i\sim\mathrm{Bernoulli}(0.99)$ (i.e.\ $1\%$ zeros) and set $b_i \leftarrow m_i\,b_i$. For directions, we drew random unit vectors $\mathbf{g}_i^{\mathrm{rand}}\sim \mathrm{Unif}(\mathbb{S}^2)$ by sampling $\tilde{\mathbf{g}}_i\sim \mathcal{N}(\mathbf{0},\mathbf{I}_3)$ and normalizing, and then sampled \texttt{bvecs} from a mixture of the typical directions $\{\mathbf{g}_i^{\mathrm{typ}}\}_{i=1}^{n}$ (i.e. even spherical grid)  and the random directions, again with weights $(0.5,0.5)$.

    \item[(ii)] \textbf{HCP-like}: We used the same procedure with $n=295$ acquisitions.
\end{itemize}

\subsubsection{B3S task configuration}
\label{sec:b3s_task_configuration}

The B3S task includes one Ball, up to three Sticks, and a Gaussian noise model (Tab.~\ref{tab:dmri_models_priors_params}). To remain comparable to \citet{manzano2024uncertainty}, we impose additional structure on this family. In particular, we tie the diffusivity across Ball and Stick compartments via a shared parameter $d$, i.e., $d_{\mathcal{B}}=d_{\mathcal{S}_1}=d_{\mathcal{S}_2}=d_{\mathcal{S}_3}=d$. Moreover only \textit{multi-shell} models were considered and we thus adapted to this case by replacing the mono-exponential nonlinearity with the appropriate multi-shell extension~\citep{jbabdi2012model}. Moreover, to mimic the strong sparsity bias used in FSL BedpostX, which employs an improper prior that penalizes non-zero fractions and thereby encourages automatic model selection \citep{behrens2007multifibre}, we use an informative prior over fractions,
$p(\mathbf{f})=\Dirichlet(\mathbf{f} \mid \boldsymbol{\alpha})$ with $\boldsymbol{\alpha}=[3.5,\,1.0,\,0.3,\,0.1]$.

Unlike \citet{manzano2024uncertainty}, we do not enforce hard angular constraints between Stick orientations. Such constraints are acquisition-specific and can become misaligned when amortizing across heterogeneous protocols; our training targets acquisition-agnostic inference (e.g., improved angular resolution with increased numbers of acquisitions), and we therefore avoid constraint sets that implicitly encode a fixed sampling scheme.

\subsubsection{BSZT and BSZT + conv task configuration}
\label{sec:bszt_task_configuration}

For the BSZT space, we include one Ball, three Sticks, three Zeppelins, and three Tensors, together with Gaussian and Rician noise models (Tab.~\ref{tab:dmri_models_priors_params}). We do not impose parameter sharing across compartments and use a uniform Dirichlet prior over fractions, $p(\mathbf{f})=\Dirichlet(\mathbf{f}\mid \boldsymbol{\alpha})$ with $\boldsymbol{\alpha}=\mathbf{1}$.

For the BSZT+conv.\ space, we further extend BSZT by adding all remaining dispersed and convolutional compartments from Table~\ref{tab:dmri_models_priors_params}, while keeping the same uniform fraction prior, $\boldsymbol{\alpha}=\mathbf{1}$.

\newpage
\renewcommand\arraystretch{0.9}
\newpage
\begin{longtable}{@{}r l l p{7.2cm} p{2.8cm}@{}}
\caption{Extended component and noise catalogue (as implemented in code).}
\label{tab:extended_symbolic_components}\\
\toprule
ID & Component & Token & Expression (implementation string) & Parameter prior \\ \midrule
\endfirsthead

\toprule
ID & Component & Token & Expression (implementation string) & Parameter prior \\ \midrule
\endhead

\midrule
\multicolumn{5}{r}{\emph{Continued on next page}}\\
\endfoot

\bottomrule
\endlastfoot

0  & Linear              & lin       & \texttt{c\_1*x} &
$\;c_1 \sim \mathcal{U}(-2,\,2)$ \\

1  & Quadratic           & quad      & \texttt{c\_1*x*x} &
$\;c_1 \sim \mathcal{U}(-0.5,\,0.5)$ \\

2  & ShiftedSquare       & shift2    & \texttt{(c\_1+x)*(c\_1+x)} &
$\;c_1 \sim \mathcal{U}(-5,\,0)$ \\

3  & Cubic               & cub       & \texttt{c\_1*x*x*x} &
$\;c_1 \sim \mathcal{U}(-0.1,\,0.1)$ \\

4  & Sinusoidal          & sin       & \texttt{c\_1*sin(c\_2*x)} &
\makecell[l]{$c_1 \sim \mathcal{U}(0,\,5)$\\$c_2 \sim \mathcal{U}(0.5,\,5)$} \\

5  & Cosinusoidal        & cos       & \texttt{c\_1*cos(c\_2*x)} &
\makecell[l]{$c_1 \sim \mathcal{U}(0,\,5)$\\$c_2 \sim \mathcal{U}(0.5,\,5)$} \\

6  & ConstantWide        & const\_w   & \texttt{c\_1} &
$\;c_1 \sim \mathcal{U}(-5,\,5)$ \\

7  & ConstantPositive    & const\_p   & \texttt{c\_1} &
$\;c_1 \sim \mathcal{U}(0,\,10)$ \\

8  & TanhRight           & tanh\_r    & \texttt{c\_1*tanh(x-c\_2)} &
\makecell[l]{$c_1 \sim \mathcal{U}(1,\,10)$\\$c_2 \sim \mathcal{U}(2,\,8)$} \\

9  & TanhLeft            & tanh\_l    & \texttt{c\_1*tanh(-x+c\_2)} &
\makecell[l]{$c_1 \sim \mathcal{U}(1,\,10)$\\$c_2 \sim \mathcal{U}(2,\,8)$} \\

10 & GaussianBump        & gauss\_b   & \texttt{c\_1*exp(-(x-c\_2)*(x-c\_2))} &
\makecell[l]{$c_1 \sim \mathcal{U}(1,\,10)$\\$c_2 \sim \mathcal{U}(2,\,8)$} \\

11 & GaussianWide        & gauss\_w   & \texttt{c\_1*exp(-(x-c\_2)*(x-c\_2)/8)} &
\makecell[l]{$c_1 \sim \mathcal{U}(1,\,10)$\\$c_2 \sim \mathcal{U}(2,\,8)$} \\

12 & RampUp              & ramp\_u    & \texttt{c\_1*Piecewise((0.0,x<c\_2),(x, x>=c\_2))} &
\makecell[l]{$c_1 \sim \mathcal{U}(1,\,5)$\\$c_2 \sim \mathcal{U}(2,\,8)$} \\

13 & RampDown            & ramp\_d    & \texttt{c\_1*Piecewise((0.0,x>c\_2),(-x+c\_2, x<=c\_2))} &
\makecell[l]{$c_1 \sim \mathcal{U}(1,\,5)$\\$c_2 \sim \mathcal{U}(2,\,8)$} \\

14 & QuarticScaled       & quart4     & \texttt{c\_1*(x/10)**4} &
$\;c_1 \sim \mathcal{U}(-5,\,5)$ \\

15 & QuinticScaled       & quint5     & \texttt{c\_1*(x/10)**5} &
$\;c_1 \sim \mathcal{U}(-5,\,5)$ \\

16 & SinusoidalPhase     & sin\_ph    & \texttt{c\_1*sin(c\_2*x + c\_3)} &
\makecell[l]{$c_1 \sim \mathcal{U}(0,\,5)$\\$c_2 \sim \mathcal{U}(0.5,\,5)$\\$c_3 \sim \mathcal{U}(-\pi,\,\pi)$} \\

17 & CosinusoidalPhase   & cos\_ph    & \texttt{c\_1*cos(c\_2*x + c\_3)} &
\makecell[l]{$c_1 \sim \mathcal{U}(0,\,5)$\\$c_2 \sim \mathcal{U}(0.5,\,5)$\\$c_3 \sim \mathcal{U}(-\pi,\,\pi)$} \\

18 & ExponentialDecay    & exp\_dec   & \texttt{c\_1*exp(-c\_2*x)} &
\makecell[l]{$c_1 \sim \mathcal{U}(0,\,10)$\\$c_2 \sim \mathcal{U}(0.1,\,2)$} \\

19 & SaturatingExponential & exp\_sat & \texttt{c\_1*(1-exp(-c\_2*x))} &
\makecell[l]{$c_1 \sim \mathcal{U}(0,\,10)$\\$c_2 \sim \mathcal{U}(0.1,\,2)$} \\

20 & Logarithmic         & log       & \texttt{c\_1*log(x+c\_2)} &
\makecell[l]{$c_1 \sim \mathcal{U}(-5,\,5)$\\$c_2 \sim \mathcal{U}(0.1,\,2)$} \\

21 & SquareRoot          & sqrt      & \texttt{c\_1*sqrt(x+c\_2)} &
\makecell[l]{$c_1 \sim \mathcal{U}(-5,\,5)$\\$c_2 \sim \mathcal{U}(0,\,2)$} \\

22 & Reciprocal          & recip     & \texttt{c\_1/(x+c\_2)} &
\makecell[l]{$c_1 \sim \mathcal{U}(-10,\,10)$\\$c_2 \sim \mathcal{U}(0.5,\,3)$} \\

23 & AbsoluteValue       & abs       & \texttt{c\_1*Abs(x-c\_2)} &
\makecell[l]{$c_1 \sim \mathcal{U}(-5,\,5)$\\$c_2 \sim \mathcal{U}(0,\,10)$} \\

24 & InverseQuadratic    & invquad   & \texttt{c\_1/(1 + c\_2*(x-c\_3)*(x-c\_3))} &
\makecell[l]{$c_1 \sim \mathcal{U}(0,\,10)$\\$c_2 \sim \mathcal{U}(0.1,\,2)$\\$c_3 \sim \mathcal{U}(0,\,10)$} \\

25 & Lorentzian          & lorentz   & \texttt{c\_1/(1 + ((x-c\_2)/c\_3)**2)} &
\makecell[l]{$c_1 \sim \mathcal{U}(0,\,10)$\\$c_2 \sim \mathcal{U}(0,\,10)$\\$c_3 \sim \mathcal{U}(0.5,\,5)$} \\

26 & Sigmoid             & sig       & \texttt{c\_1/(1+exp(-c\_2*(x-c\_3)))} &
\makecell[l]{$c_1 \sim \mathcal{U}(0,\,10)$\\$c_2 \sim \mathcal{U}(0.1,\,5)$\\$c_3 \sim \mathcal{U}(0,\,10)$} \\

27 & DampedSinusoidal    & d\_sin     & \texttt{c\_1*exp(-c\_2*x)*sin(c\_3*x)} &
\makecell[l]{$c_1 \sim \mathcal{U}(0,\,5)$\\$c_2 \sim \mathcal{U}(0.05,\,1)$\\$c_3 \sim \mathcal{U}(0.5,\,8)$} \\

28 & DampedCosinusoidal  & d\_cos     & \texttt{c\_1*exp(-c\_2*x)*cos(c\_3*x)} &
\makecell[l]{$c_1 \sim \mathcal{U}(0,\,5)$\\$c_2 \sim \mathcal{U}(0.05,\,1)$\\$c_3 \sim \mathcal{U}(0.5,\,8)$} \\

29 & TanhCentered        & tanh\_c    & \texttt{c\_1*tanh(c\_2*(x-c\_3))} &
\makecell[l]{$c_1 \sim \mathcal{U}(-10,\,10)$\\$c_2 \sim \mathcal{U}(0.1,\,2)$\\$c_3 \sim \mathcal{U}(2,\,8)$} \\

30 & ExponentialGrowth   & exp\_grow  & \texttt{c\_1*exp(c\_2*x)} &
\makecell[l]{$c_1 \sim \mathcal{U}(0,\,10)$\\$c_2 \sim \mathcal{U}(0.05,\,0.8)$} \\

31 & PowerLawDecay       & pow\_dec   & \texttt{c\_1/(x + c\_2)**c\_3} &
\makecell[l]{$c_1 \sim \mathcal{U}(0,\,10)$\\$c_2 \sim \mathcal{U}(0.5,\,5)$\\$c_3 \sim \mathcal{U}(0.5,\,3)$} \\

32 & ArctangentStep      & atan      & \texttt{c\_1*atan(c\_2*(x-c\_3))} &
\makecell[l]{$c_1 \sim \mathcal{U}(0,\,10)$\\$c_2 \sim \mathcal{U}(0.1,\,2)$\\$c_3 \sim \mathcal{U}(0,\,10)$} \\

33 & HyperbolicSecant    & sech      & \texttt{c\_1*sech(c\_2*(x-c\_3))} &
\makecell[l]{$c_1 \sim \mathcal{U}(0,\,8)$\\$c_2 \sim \mathcal{U}(0.1,\,2)$\\$c_3 \sim \mathcal{U}(0,\,10)$} \\

34 & SincDecay           & sinc      & \texttt{c\_1*sin(c\_2*x)/(x+c\_3)} &
\makecell[l]{$c_1 \sim \mathcal{U}(0,\,5)$\\$c_2 \sim \mathcal{U}(0.5,\,5)$\\$c_3 \sim \mathcal{U}(0.5,\,5)$} \\

35 & AbsoluteSinusoidal  & abs\_sin   & \texttt{c\_1*Abs(sin(c\_2*x + c\_3))} &
\makecell[l]{$c_1 \sim \mathcal{U}(0,\,5)$\\$c_2 \sim \mathcal{U}(0.5,\,5)$\\$c_3 \sim \mathcal{U}(-\pi,\,\pi)$} \\

36 & RectifiedLinear     & relu      & \texttt{c\_1*Piecewise((0, x < c\_2), (x - c\_2, True))} &
\makecell[l]{$c_1 \sim \mathcal{U}(0,\,5)$\\$c_2 \sim \mathcal{U}(0,\,8)$} \\

37 & Softplus            & softplus  & \texttt{c\_1*log(1+exp(c\_2*(x-c\_3)))/c\_2} &
\makecell[l]{$c_1 \sim \mathcal{U}(0,\,5)$\\$c_2 \sim \mathcal{U}(0.1,\,5)$\\$c_3 \sim \mathcal{U}(0,\,10)$} \\

38 & Gompertz            & gomp      & \texttt{c\_1*exp(-c\_2*exp(-c\_3*x))} &
\makecell[l]{$c_1 \sim \mathcal{U}(0,\,10)$\\$c_2 \sim \mathcal{U}(0.1,\,3)$\\$c_3 \sim \mathcal{U}(0.05,\,1.5)$} \\

39 & LinearFractional    & linfrac   & \texttt{c\_1*x/(1 + c\_2*x)} &
\makecell[l]{$c_1 \sim \mathcal{U}(-5,\,5)$\\$c_2 \sim \mathcal{U}(0.1,\,2)$} \\

40 & SineSquared         & sin2      & \texttt{c\_1*sin(c\_2*x)**2} &
\makecell[l]{$c_1 \sim \mathcal{U}(0,\,5)$\\$c_2 \sim \mathcal{U}(0.5,\,5)$} \\

41 & TriangularBump      & tri       & \texttt{c\_1*Piecewise((0, Abs((x-c\_2)/c\_3) >= 1), (1 - Abs((x-c\_2)/c\_3), True))} &
\makecell[l]{$c_1 \sim \mathcal{U}(0,\,5)$\\$c_2 \sim \mathcal{U}(0.5,\,5)$\\$c_3 \sim \mathcal{U}(0.5,\,5)$} \\

42 & NoiseObserver            & n\_obs  & \texttt{normal(c\_1)} &
$\;c_1 \sim \mathcal{U}(0.1,\,2)$ \\

43 & NoiseIncreasing          & n\_inc  & \texttt{normal(c\_1) * (x+1)} &
$\;c_1 \sim \mathcal{U}(0.5,\,2)$ \\

44 & NoiseDecreasing          & n\_dec  & \texttt{normal(c\_1) * (11 - x)} &
$\;c_1 \sim \mathcal{U}(0.5,\,2)$ \\

45 & NoiseQuadratic           & n\_quad & \texttt{normal(c\_1) *(x**2 + 1)} &
$\;c_1 \sim \mathcal{U}(0.2,\,1)$ \\

46 & NoiseQuadraticDecreasing & n\_qdec & \texttt{normal(c\_1) * (11 - x**2)} &
$\;c_1 \sim \mathcal{U}(0.2,\,1)$ \\

47 & NoiseExponential         & n\_exp  & \texttt{normal(c\_1) * exp(c\_2 * x)} &
\makecell[l]{$c_1 \sim \mathcal{U}(0,\,5)$\\$c_2 \sim \mathcal{U}(0.05,\,0.5)$} \\

48 & NoiseSigmoid             & n\_sig  & \texttt{normal(c\_1)/(1+exp(-c\_2*(x-c\_3)))} &
\makecell[l]{$c_1 \sim \mathcal{U}(0,\,5)$\\$c_2 \sim \mathcal{U}(0.1,\,1)$\\$c_3 \sim \mathcal{U}(0,\,10)$} \\

49 & NoisePeaked              & n\_peak & \texttt{normal(c\_1) * exp(-((x-c\_2)**2)/(c\_3 + 1e-3))} &
\makecell[l]{$c_1 \sim \mathcal{U}(0,\,5)$\\$c_2 \sim \mathcal{U}(0,\,10)$\\$c_3 \sim \mathcal{U}(0.5,\,5)$} \\

\end{longtable}

\newpage
\begin{longtable}{@{}r l l p{6.8cm} p{.5cm} p{3.5cm}@{}}
\caption{Diffusion MRI signal models (\texttt{DMIPY}-style) with per-parameter priors.}
\label{tab:dmri_models_priors_params}\\
\toprule
\textbf{ID} & \textbf{Name} & \textbf{Symbol} & \textbf{Signal model } & \textbf{$\vtheta$} & \textbf{Prior} \\
\midrule
\endfirsthead

\toprule
\textbf{ID} & \textbf{Name} & \textbf{Symbol} & \textbf{Signal model } & \textbf{$\vtheta$} & \textbf{Prior} \\
\midrule
\endhead

\midrule
\multicolumn{6}{r}{\emph{Continued on next page}}\\
\endfoot

\bottomrule
\endlastfoot

\multirow{1}{*}{0} &
\multirow{1}{*}{Ball} &
\multirow{1}{*}{B} &
\multirow{1}{=}{\(\displaystyle S_B(b)=\exp\!\big(-b\,d_{\mathrm{iso}}\big)\)} &
\(d_{\mathrm{iso}}\) &
\(\displaystyle d_{\mathrm{iso}}\sim \Uniform\!\big(0,0.01\big)\) \\
\\[-1.0ex]

\multirow{2}{*}{1} &
\multirow{2}{*}{Stick} &
\multirow{2}{*}{S} &
\multirow{2}{=}{\(\displaystyle S_S(b,\vec{\mathbf{b}})=\exp\!\left(-b\,d_{\parallel}(\vec{\mathbf{b}}\cdot\mathbf{n})^2\right)\)} &
\(d_{\parallel}\) &
\(\displaystyle d_{\parallel}\sim \Uniform\!\big(0,\;0.01\big)\ \) \\
& & & & \(\mathbf{n}\) &
\(\displaystyle \mathbf{n}\sim \Uniform(\mathbb{S}^2_+)\) \\
\\[-1.0ex]

\multirow{3}{*}{2} &
\multirow{3}{*}{Zeppelin} &
\multirow{3}{*}{Z} &
\multirow{3}{=}{\(\displaystyle
S_Z(b,\vec{\mathbf{b}})=\exp\!\left(-b\left[d_{\parallel}(\vec{\mathbf{b}}\cdot\mathbf{n})^2+d_{\perp}\big(1-(\vec{\mathbf{b}}\cdot\mathbf{n})^2\big)\right]\right)
\)} &
\(d_{\parallel}\) &
\(\displaystyle d_{\parallel}\sim \Uniform\!\big(0,\;0.01\big) \) \\
& & & & \(d_{\perp}\) &
\(\displaystyle d_{\perp}\sim \Uniform(0,0.01)\) \\
& & & & \(\mathbf{n}\) &
\(\displaystyle \mathbf{n}\sim \Uniform(\mathbb{S}^2_+)\) \\
\\[-1.0ex]

\multirow{4}{*}{3} &
\multirow{4}{*}{Tensor (DTI)} &
\multirow{4}{*}{T} &
\multirow{4}{=}{\(\displaystyle
S_T(b,\vec{\mathbf{b}})=\exp\!\left(-b\,\vec{\mathbf{b}}^{\top}\mathbf{D}\,\vec{\mathbf{b}}\right),
\quad \mathbf{D}\succ 0
\)}
& \(\boldsymbol{D}\) &
\(\displaystyle \boldsymbol{\theta}\sim \mathcal{N}(\mathbf{0},\,\mathbf{I}_6)\) \\

& & & &  &
\(\displaystyle \mathbf{L}=\mathrm{tril}(\boldsymbol{\theta}) \in \mathbb{R}^{3\times 3}\) \\

& & & &  &
\(\displaystyle \tilde{\mathbf{D}}=\mathbf{L}\mathbf{L}^{\top}\) \\

& & & &  \(\) &
\(\displaystyle \mathbf{D}=D_{\mathrm{scale}}\,\tilde{\mathbf{D}}+\lambda_{\min}\mathbf{I}_3\) \\

& & & &  &
\(\displaystyle D_{\text{scale}}=10^{-3},\lambda_{\text{min}} = 10^{-4} \) \\[1.0ex]

\multirow{3}{*}{4} &
\multirow{3}{*}{Watson Stick} &
\multirow{3}{*}{S$_\mathcal{W}$} &
\multirow{3}{=}{\(\displaystyle
S_{WS}(b,\vec{\mathbf{b}})=\int_{\mathbb{S}^2} S_S(b,\vec{\mathbf{b}}\mid \mathbf{n}, d_\parallel)\, f_\mathcal{W}(\mathbf{n}\mid \boldsymbol{\mu},\kappa)\,d\mathbf{n}
\)} &
\(d_{\parallel}\) &
\(\displaystyle d_{\parallel}\sim \Uniform\!\big(0,\;0.01\big)\) \\
& & & & \(\boldsymbol{\mu}\) &
\(\displaystyle \boldsymbol{\mu}\sim \mathrm{Unif}(\mathbb{S}^2_+)\) \\

& & & & \(\mathrm{ODI}\) &
\(\displaystyle \mathrm{ODI}\sim \mathrm{Unif}(0,1)\) \\

& & & &  &
\(\displaystyle \kappa \;=\; \frac{1}{\tan\!\left(\mathrm{ODI}\,\frac{\pi}{2}\right)}\) \\
\\[-1.0ex]

\multirow{5}{*}{5} &
\multirow{5}{*}{Bingham Stick} &
\multirow{5}{*}{S$_\mathcal{B}$} &
\multirow{5}{=}{\(\displaystyle
S_{BS}(b,\vec{\mathbf{b}})=\int_{\mathbb{S}^2} S_S(b,\vec{\mathbf{b}}\mid \mathbf{n})\, f_B(\mathbf{n}\mid \boldsymbol{\mu},\kappa,\beta,\psi)\,d\mathbf{n}
\)} &
\(d_{\parallel}\) &
\(\displaystyle d_{\parallel}\sim \Uniform\!\big(0,\;0.01\big)\) \\
& & & & \(\boldsymbol{\mu}\) &
\(\displaystyle \boldsymbol{\mu}\sim \Uniform(\mathbb{S}^2_+)\) \\
& & & & \(\mathrm{ODI}\) &
\(\displaystyle \mathrm{ODI}\sim \Uniform(0,1)\) \\
& & & & \(\beta_{\mathrm{frac}}\) &
\(\displaystyle \beta_{\mathrm{frac}}\sim \Uniform(0,1)\) \\
& & & & \(\psi\) &
\(\displaystyle \psi\sim \Uniform(0,\pi)\) \\
\\
& & & &  &
\(\displaystyle \kappa \;=\; \frac{1}{\tan\!\left(\mathrm{ODI}\,\frac{\pi}{2}\right)}\) \\
\\
& & & &  &
\(\displaystyle \beta \;=\; \beta_{\text{frac}} \kappa \) \\
\\[-1.0ex]
\multirow{5}{*}{6} &
\multirow{5}{*}{Watson Zeppelin} &
\multirow{5}{*}{Z$_\mathcal{W}$} &
\multirow{5}{=}{\(\displaystyle
S_{WZ}(b,\vec{\mathbf{b}})=\int_{\mathbb{S}^2} S_Z(b,\vec{\mathbf{b}}\mid \mathbf{n})\, f_\mathcal{W}(\mathbf{n}\mid \boldsymbol{\mu},\kappa)\,d\mathbf{n}
\)} &
\(d_{\parallel}\) &
\(\displaystyle d_{\parallel}\sim \Uniform\!\big(0,\;3\times10^{-9}\big)\) \\

& & & & \(d_{\perp}\) &
\(\displaystyle d_{\perp}\sim \Uniform(0,\;d_{\parallel})\) \\

& & & & \(\boldsymbol{\mu}\) &
\(\displaystyle \boldsymbol{\mu}\sim \mathcal{U}(\mathbb{S}^2_+)\) \\

& & & & \(\mathrm{ODI}\) &
\(\displaystyle \mathrm{ODI}\sim \mathcal{U}\!\big(0.,\;1.\big)\) \\

& & & &  &
\(\displaystyle \kappa \;=\; \frac{1}{\tan\!\left(\mathrm{ODI}\,\frac{\pi}{2}\right)}\) \\
\\[-1.0ex]

\multirow{8}{*}{7} &
\multirow{8}{*}{Bingham Zeppelin} &
\multirow{8}{*}{Z$_\mathcal{B}$} &
\multirow{8}{=}{\(\displaystyle
S_{BZ}(b,\vec{\mathbf{b}})=\int_{\mathbb{S}^2} S_Z(b,\vec{\mathbf{b}}\mid \mathbf{n})\, f_\mathcal{B}(\mathbf{n}\mid \boldsymbol{\mu},\kappa,\beta,\psi)\,d\mathbf{n}
\)} &
\(d_{\parallel}\) &
\(\displaystyle d_{\parallel}\sim \Uniform\!\big(0,\;0.01\big)\) \\

& & & & \(d_{\perp}\) &
\(\displaystyle d_{\perp}\sim \Uniform(0,\;0.01)\) \\

& & & & \(\boldsymbol{\mu}\) &
\(\displaystyle \boldsymbol{\mu}\sim \mathcal{U}(\mathbb{S}^2_+)\) \\

& & & & \(\mathrm{ODI}\) &
\(\displaystyle \mathrm{ODI}\sim \mathcal{U}\!\big(0.,\;1.\big)\) \\

& & & & \(\beta_{\mathrm{frac}}\) &
\(\displaystyle \beta_{\mathrm{frac}}\sim \mathcal{U}\!\big(0,\;1\big)\) \\

& & & & \(\psi\) &
\(\displaystyle \psi\sim \mathcal{U}\!\big(0.,\;\pi\big)\) \\

& & & &  &
\(\displaystyle \kappa \;=\; \frac{1}{\tan\!\left(\mathrm{ODI}\,\frac{\pi}{2}\right)}\) \\

& & & &  &
\(\displaystyle \beta \;=\; \beta_{\mathrm{frac}}\,\kappa\) \\
\\[-1.0ex]
\multirow{6}{*}{9} &
\multirow{6}{*}{NODDI (Watson)} &
\multirow{6}{*}{No$_\mathcal{W}$} &
\multirow{6}{=}{\(\displaystyle
S_{\mathrm{NODDI}\_\mathcal{W}}(b,\vec{\mathbf{b}})
=
f\,S_{WS}(b,\vec{\mathbf{b}};\boldsymbol{\mu},\mathrm{ODI},\lambda_{\parallel})
+(1-f)\,S_{WZ}(b,\vec{\mathbf{b}};\boldsymbol{\mu},\mathrm{ODI},\lambda_{\perp},\lambda_{\parallel})
\)} &
\(f\) &
\(\displaystyle f\sim \mathcal{U}\!\big(0.,\;1.\big)\) \\

& & & & \(\lambda_{\perp}\) &
\(\displaystyle \lambda_{\perp}\sim \mathcal{U}\!\big(0.,\;0.01\big)\) \\

& & & & \(\lambda_{\parallel}\) &
\(\displaystyle \lambda_{\parallel}\sim \mathcal{U}\!\big(0.,\;0.01\big)\) \\

& & & & \(\boldsymbol{\mu}\) &
\(\displaystyle \boldsymbol{\mu}\sim \mathcal{U}(\mathbb{S}^2_+)\) \\

& & & & \(\mathrm{ODI}\) &
\(\displaystyle \mathrm{ODI}\sim \mathcal{U}\!\big(0.,\;1.\big)\) \\

& & & &  &
\(\displaystyle \kappa \;=\; \frac{1}{\tan\!\left(\mathrm{ODI}\,\frac{\pi}{2}\right)}\) \\
\\[-1.0ex]

\multirow{8}{*}{10} &
\multirow{8}{*}{NODDI (Bingham)} &
\multirow{8}{*}{No$_\mathcal{B}$} &
\multirow{8}{=}{\(\displaystyle
S_{\mathrm{NODDI}\_\mathcal{B}}(b,\vec{\mathbf{b}})
=
f\,S_{BS}(b,\vec{\mathbf{b}};\boldsymbol{\mu},\mathrm{ODI},\beta_{\mathrm{frac}},\psi,\lambda_{\parallel})
+(1-f)\,S_{BZ}(b,\vec{\mathbf{b}};\boldsymbol{\mu},\mathrm{ODI},\beta_{\mathrm{frac}},\psi,\lambda_{\perp},\lambda_{\parallel})
\)} &
\(f\) &
\(\displaystyle f\sim \mathcal{U}\!\big(0.,\;1.\big)\) \\

& & & & \(\lambda_{\perp}\) &
\(\displaystyle \lambda_{\perp}\sim \mathcal{U}\!\big(0.,\;0.01\big)\) \\

& & & & \(\lambda_{\parallel}\) &
\(\displaystyle \lambda_{\parallel}\sim \mathcal{U}\!\big(0.,\;0.01\big)\) \\

& & & & \(\boldsymbol{\mu}\) &
\(\displaystyle \boldsymbol{\mu}\sim \mathcal{U}(\mathbb{S}^2_+)\) \\

& & & & \(\mathrm{ODI}\) &
\(\displaystyle \mathrm{ODI}\sim \mathcal{U}\!\big(0.,\;1.\big)\) \\

& & & & \(\beta_{\mathrm{frac}}\) &
\(\displaystyle \beta_{\mathrm{frac}}\sim \mathcal{U}\!\big(0.,\;1.\big)\) \\

& & & & \(\psi\) &
\(\displaystyle \psi\sim \mathcal{U}\!\big(0.,\;\pi\big)\) \\

& & & &  &
\(\displaystyle \kappa \;=\; \frac{1}{\tan\!\left(\mathrm{ODI}\,\frac{\pi}{2}\right)},\qquad
\beta \;=\; \beta_{\mathrm{frac}}\,\kappa\) \\
\\[-1.0ex]

\multirow{8}{*}{9} &
\multirow{8}{*}{SANDI (Watson)} &
\multirow{8}{*}{Sa$_\mathcal{W}$} &
\multirow{8}{=}{\(\displaystyle
S_{\mathrm{SANDI}\_\mathcal{W}}(b,\vec{\mathbf{b}})
=
(1-f_{\mathrm{ec}})\Big[
f_{\mathrm{in}}\,S_{WS}\!\big(b,\vec{\mathbf{b}};\boldsymbol{\mu},\mathrm{ODI},\lambda_{\parallel}^{\mathrm{in}}\big)
+(1-f_{\mathrm{in}})\cdot 1
\Big]
+
f_{\mathrm{ec}}\,S_{WZ}\!\big(b,\vec{\mathbf{b}};\boldsymbol{\mu},\mathrm{ODI},\lambda_{\perp}^{\mathrm{ex}},\lambda_{\parallel}^{\mathrm{ex}}\big)
\)} &
\(f_{\mathrm{in}}\) &
\(\displaystyle f_{\mathrm{in}}\sim \mathcal{U}\!\big(0.,\;1.\big)\) \\

& & & & \(f_{\mathrm{ec}}\) &
\(\displaystyle f_{\mathrm{ec}}\sim \mathcal{U}\!\big(0.,\;1.\big)\) \\

& & & & \(\lambda_{\parallel}^{\mathrm{in}}\) &
\(\displaystyle \lambda_{\parallel}^{\mathrm{in}}\sim \mathcal{U}\!\big(0.,\;0.01\big)\) \\

& & & & \(\lambda_{\perp}^{\mathrm{ex}}\) &
\(\displaystyle \lambda_{\perp}^{\mathrm{ex}}\sim \mathcal{U}\!\big(0.,\;0.01\big)\) \\

& & & & \(\lambda_{\parallel}^{\mathrm{ex}}\) &
\(\displaystyle \lambda_{\parallel}^{\mathrm{ex}}\sim \mathcal{U}\!\big(0.,\;0.01\big)\) \\

& & & & \(\boldsymbol{\mu}\) &
\(\displaystyle \boldsymbol{\mu}\sim \mathcal{U}(\mathbb{S}^2_+)\) \\

& & & & \(\mathrm{ODI}\) &
\(\displaystyle \mathrm{ODI}\sim \mathcal{U}\!\big(0.,\;1.\big)\) \\

& & & &  &
\(\displaystyle \kappa \;=\; \frac{1}{\tan\!\left(\mathrm{ODI}\,\frac{\pi}{2}\right)}\) \\
\\[-1.0ex]

\multirow{11}{*}{10} &
\multirow{11}{*}{SANDI (Bingham)} &
\multirow{11}{*}{Sa$_\mathcal{B}$} &
\multirow{11}{=}{\(\displaystyle
S_{\mathrm{SANDI}\_\mathcal{B}}(b,\vec{\mathbf{b}})
=
(1-f_{\mathrm{ec}})\Big[
f_{\mathrm{in}}\,S_{BS}\!\big(b,\vec{\mathbf{b}};\boldsymbol{\mu},\mathrm{ODI},\beta_{\mathrm{frac}},\psi,\lambda_{\parallel}^{\mathrm{in}}\big)
+(1-f_{\mathrm{in}})\cdot 1
\Big]
+
f_{\mathrm{ec}}\,S_{BZ}\!\big(b,\vec{\mathbf{b}};\boldsymbol{\mu},\mathrm{ODI},\beta_{\mathrm{frac}},\psi,\lambda_{\perp}^{\mathrm{ex}},\lambda_{\parallel}^{\mathrm{ex}}\big)
\)} &
\(f_{\mathrm{in}}\) &
\(\displaystyle f_{\mathrm{in}}\sim \mathcal{U}\!\big(0.,\;1.\big)\) \\

& & & & \(f_{\mathrm{ec}}\) &
\(\displaystyle f_{\mathrm{ec}}\sim \mathcal{U}\!\big(0.,\;1.\big)\) \\

& & & & \(\lambda_{\parallel}^{\mathrm{in}}\) &
\(\displaystyle \lambda_{\parallel}^{\mathrm{in}}\sim \mathcal{U}\!\big(0.,\;0.01\big)\) \\

& & & & \(\lambda_{\perp}^{\mathrm{ex}}\) &
\(\displaystyle \lambda_{\perp}^{\mathrm{ex}}\sim \mathcal{U}\!\big(0.,\;0.01\big)\) \\

& & & & \(\lambda_{\parallel}^{\mathrm{ex}}\) &
\(\displaystyle \lambda_{\parallel}^{\mathrm{ex}}\sim \mathcal{U}\!\big(0.,\;0.01\big)\) \\

& & & & \(\boldsymbol{\mu}\) &
\(\displaystyle \boldsymbol{\mu}\sim \mathcal{U}(\mathbb{S}^2_+)\) \\

& & & & \(\mathrm{ODI}\) &
\(\displaystyle \mathrm{ODI}\sim \mathcal{U}\!\big(0.,\;1.\big)\) \\

& & & & \(\beta_{\mathrm{frac}}\) &
\(\displaystyle \beta_{\mathrm{frac}}\sim \mathcal{U}\!\big(0.,\;1.\big)\) \\

& & & & \(\psi\) &
\(\displaystyle \psi\sim \mathcal{U}\!\big(0.,\;\pi\big)\) \\

& & & &  &
\(\displaystyle \kappa \;=\; \frac{1}{\tan\!\left(\mathrm{ODI}\,\frac{\pi}{2}\right)}\) \\

& & & &  &
\(\displaystyle \beta \;=\; \beta_{\mathrm{frac}}\,\kappa\) \\
\\[-1.0ex]

\multirow{2}{*}{10} &
\multirow{2}{*}{Gaussian noise} &
\multirow{2}{*}{G} &
\multirow{2}{=}{\(\displaystyle y = S_0\,S(\cdot) + \epsilon,\quad \epsilon\sim\mathcal{N}(0,\sigma^2)\)} &
\(\sigma\) &
\(\displaystyle \mathrm{SNR}\sim \mathcal{U}(3,\;80)\) \\

& & & &  &
\(\displaystyle \sigma = \frac{1}{\mathrm{SNR}}\) \\
\\[-1.0ex]

\multirow{2}{*}{11} &
\multirow{2}{*}{Rician noise} &
\multirow{2}{*}{R} &
\multirow{2}{=}{\(\displaystyle
y \sim \mathrm{Rice}(\nu=S_0S(\cdot),\sigma),\
p(y\mid\nu,\sigma)=\frac{y}{\sigma^2}\exp\!\Big(-\frac{y^2+\nu^2}{2\sigma^2}\Big)\,
I_0\!\Big(\frac{y\nu}{\sigma^2}\Big)
\)} &
\(\sigma\) &
\(\displaystyle \mathrm{SNR}\sim \mathcal{U}(3,\;80)\) \\

& & & & &
\(\displaystyle \sigma = \frac{1}{\mathrm{SNR}}\) \\
\vspace{1cm}

\end{longtable}

\end{document}